\documentclass[journal]{IEEEtran}
\usepackage{amsmath,amsfonts}
\usepackage{algorithmic}
\usepackage{array}
\usepackage{textcomp}
\usepackage{stfloats}
\usepackage{url}
\usepackage{verbatim}
\usepackage{graphicx}

\usepackage{lineno,hyperref}
\modulolinenumbers[5]
\usepackage{amssymb}
\usepackage{amsfonts}
\usepackage{amsmath}
\usepackage{subfigure}
\usepackage{color}
\usepackage{blindtext}
\usepackage{algorithm}
\usepackage{algorithmic}
\usepackage{cite}
\usepackage{amsthm}
\usepackage[normalem]{ulem}
\usepackage{url}
\usepackage{comment}
\usepackage{multirow}
\usepackage{booktabs}
\usepackage{diagbox}
\usepackage{xcolor}
\usepackage{diagbox}
\usepackage{indentfirst}
\usepackage{pifont}
\usepackage{bm}

\renewcommand\arraystretch{1.3}

\newtheorem{Def}{Definition}
\usepackage[normalem]{ulem}
\newcounter{ToDo}
\newcounter{gaocomm} 
\newcounter{Note}
\definecolor{blue-violet}{rgb}{0.00,0.75,0.90}
\definecolor{mygreen}{rgb}{0.0, 0.5, 0.0}
\definecolor{awesome}{rgb}{1.0, 0.13, 0.32}
\definecolor{bostonuniversityred}{rgb}{1.0, 0.0, 0.0}
\def\BibTeX{{\rm B\kern-.05em{\sc i\kern-.025em b}\kern-.08em
    T\kern-.1667em\lower.7ex\hbox{E}\kern-.125emX}}
\usepackage{balance}

\begin{document}
\title{DGNN: Decoupled Graph Neural Networks with Structural Consistency between Attribute and Graph Embedding Representations}
\author{Jinlu Wang,
        Jipeng Guo,
        Yanfeng Sun, 
        Junbin Gao,
        Shaofan Wang,
        Yachao Yang,
        and Baocai Yin, \IEEEmembership{Member,~IEEE}
\thanks{Jinlu Wang, Yanfeng Sun, Shaofan Wang, Yachao Yang and Baocai Yin are with Beijing Key Laboratory of Multimedia and Intelligent Software Technology, Faculty of Information Technology, Beijing University of Technology, Beijing 100124, China. E-mail: \{wangjinlu, yangyc\}@emails.bjut.edu.cn, \{yfsun, wangshaofan, ybc\}@bjut.edu.cn}
\thanks{Jipeng Guo is with College of Information Science and Technology, Beijing University of Chemical Technology, Beijing 100029, China. \protect E-mail: guojipeng@buct.edu.cn}
\thanks{Junbin Gao is with the Discipline of Business Analytics, The University of Sydney Business School, The University of Sydney, Camperdown, NSW 2006, Australia. \protect E-mail: junbin.gao@sydney.edu.au}
\thanks{Corresponding authors: Jipeng Guo and Yanfeng Sun}
%\thanks{Manuscript received January 1, 2015; revised December XX, 2015.}
}

\markboth{Journal of \LaTeX\ Class Files,~Vol.~18, No.~9, September~2020}%
{DGNN: Decoupled Graph Neural Networks with Structural Consistency between Attribute and Graph Embedding Representations}

\maketitle

\begin{abstract}
Graph neural networks (GNNs) demonstrate a robust capability for representation learning on graphs with complex structures, showcasing superior performance in various applications. The majority of existing GNNs employ a graph convolution operation by using both attribute and structure information through coupled learning. In essence, GNNs, from an optimization perspective, seek to learn a consensus and compromise embedding representation that balances attribute and graph information, selectively exploring and retaining valid information. To obtain a more comprehensive embedding representation of nodes, a novel GNNs framework, dubbed Decoupled Graph Neural Networks (DGNN), is introduced. DGNN explores distinctive embedding representations from the attribute and graph spaces by decoupled terms. Considering that semantic graph, constructed from attribute feature space, consists of different node connection information and provides enhancement for the topological graph, both topological and semantic graphs are combined for the embedding representation learning. Further, structural consistency among attribute embedding and graph embeddings is promoted to effectively remove redundant information and establish soft connection. This involves promoting factor sharing for adjacency reconstruction matrices, facilitating the exploration of  a consensus and
high-level correlation. Finally, a more powerful and complete representation is achieved through the concatenation of these embeddings. Experimental results conducted on several graph benchmark datasets verify its superiority in node classification task.

\end{abstract}


\section{Introduction}
\IEEEPARstart{G}{raphs} are widespread in the real word, such as social networks, knowledge graph, and biological networks, etc \cite{zhang2019graph}. Typically, the graph data encapsulate both node attributes and topology structure as characteristics. It is imperative and highly significant to derive discriminative embedding representations for complex graph data. Recently, graph
neural networks (GNNs) have demonstrated powerful capability in handling irregular data, playing a pivotal role in enhancing the performance in many down-stream tasks such as node classification \cite{kipf2017semi,xie2022active,xie2023semisupervised}, traffic forecasting \cite{jiang2022graph,zheng2023dstagcn}, link prediction \cite{zhang2018link,cai2022line}, and node clustering \cite{wang2023overview,xiao2023sgae,zhao2022adaptive}, etc. Furthermore, the GNNs have been promoted in various fields, including recommendation systems \cite{wu2021comprehensive, zhou2020graph}, natural
language processing \cite{wu2023graph}, fraud detection \cite{li2023lgm, jiang2021mafi} and information retrieval \cite{li2020quaternion}. In a nutshell, with the the message passing mechanism, GNNs generate the embedding representations of a target node by transforming and then iteratively aggregating the representations of its neighbors \cite{wu2021comprehensive}. The resultant graph embedding representation is inherently discriminative, encapsulating both attribute feature and structural information. In essence, various GNNs methods utilize different feature aggregation operators to aggregate neighbor information.

The existing GNNs methods can be broadly categorized into two main families: spectral GNNs methods \cite{kipf2017semi, wu2019simplifying,defferrard2016convolutional, chien2020adaptive, he2021bernnet,levie2018cayleynets} and spatial GNNs methods \cite{xu2018powerful, velickovic2017graph, huang2020combining, hamilton2017inductive}. Inspired by the spectral graph theory \cite{chung1997spectral}, the spectral GNNs methods leverage graph Fourier
transformation to learn node embedding representations. For instance,, Bruna et al. \cite{bruna2013spectral} utilize the Fourier basis of spectral graph to extend Euclidean convolution operation into non-Euclidean graph data. Defferrard et al.  \cite{defferrard2016convolutional} adopt the Chebyshev polynomials filter as simple convolution and propose the ChebNet. Further, Kipf et al. \cite{kipf2017semi} simplify  ChebNet by utilizing the first-order approximation of Chebyshev polynomials, resulting in the development of the Graph Convolutional Network (GCN). In the context where nonlinear transformation is deemed unnecessary, Wu et al. \cite{wu2019simplifying} propose the Simplifying Graph Convolutional networks (SGC), incorporating  linear transformation weight matrix. The core idea behind  GCN is to apply a low-pass filter to node signal, smoothing the connected nodes and obtaining similar representations. In contrast to spectral GNNs, the spatial methods \cite{xu2018powerful,velickovic2017graph,huang2020combining} define graph convolutions by aggregating and transforming local adjacency information. Hamilton et al. \cite{hamilton2017inductive}  introduce the concept of sampling and aggregating neighbors to target node, utlizing learnable aggregators. To distinguish the importance of each neighbor, Graph Attention Network
(GAT) \cite{velickovic2017graph} is proposed to adaptively learn attention scores for neighbors. Further details and surveys on various GNN methods can be found in the literature. Among these methods, GCN  stands out as one of the most representative due to its simplicity and effectiveness.

However, the stacking of multiple layers in GCN often leads to the over-smoothing issue \cite{li2018deeper,yang2019dual}. To address this challenge, several methods focus on enhancing the message propagation mechanism. Examples include Personalized Propagation of Neural Predictions (PPNP) \cite{gasteiger2018predict} and its approximation (APPNP) \cite{gasteiger2018predict}, which aims to reduce the high computational complexity. Other approaches, such as Jumping Knowledge Network (JKNet)\cite{xu2018representation}, Graph Convolutional Network via Initial residual and Identity mapping (GCNII)\cite{chen2020simple}, and Deep Adaptive Graph
Neural Network (DAGNN) \cite{liu2020towards}, also focus on mitigating the over-smoothing challenge.
JKNet and DAGNN try to utilize cross-scale information from various layers to alleviate the over-smoothing issue. On the other hand, APPNP and GCNII improve the importance of the node itself by establishing connection between input and output of convolutional layer. These approaches improve upon the classic GCN by modifying graph convolution operator or redesigning message aggregation mechanisms,  resulting in  excellent performances. 

\begin{figure*}
\centering
\includegraphics[width=17cm]{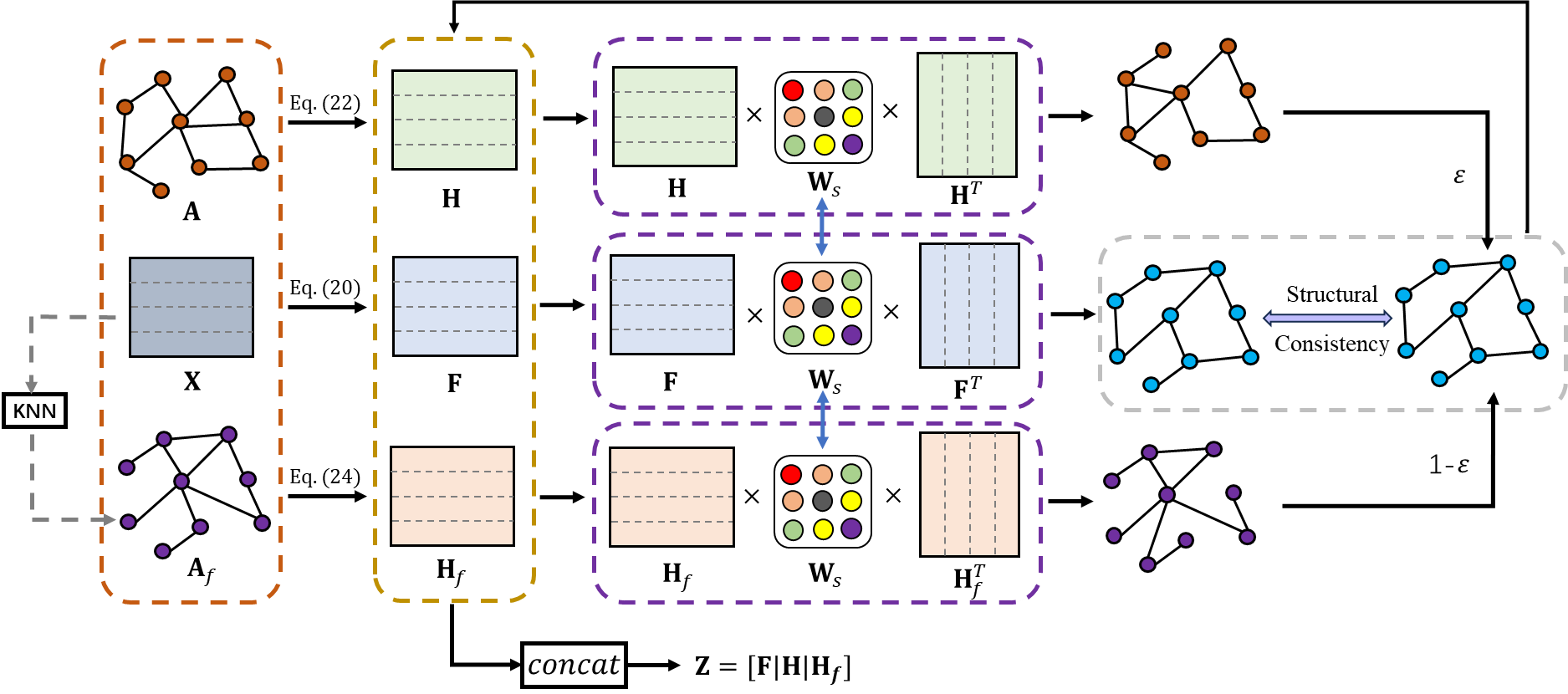}
\caption{The overall framework of DGNN model, which  simultaneously learns embedding representations from node attribute, 
topological graph, and semantic graph. The structural consistency aspect adeptly establishes the correlation between different representations,  eliminating feature redundancy and achieving structural consistency.}\label{framework}
\end{figure*} 

To enhance GNNs more effectively, several unified optimization frameworks have been proposed to understand various GCNs
and reveal principles of aggregation operations from a uniform perspective. Ma et al. \cite{ma2021unified} summarize several representative GNNs methods as a graph signal denoising problem with different
Laplacian regularizers and propose the adaptive local smoothing GNNs. Considering that the core objective of GNNs is to encode discriminative information
from attribute and smooth similar nodes, Zhu et al. \cite{zhu2021interpreting} have designed a unified optimization framework. This framework incorporates a feature fitting function over graph kernels along with a graph regularization term. They formalize the propagation mechanisms of various GNNs as optimal solution within this framework. Further, Wang et al. \cite{wang2023beyond} have introduced a regularizer-induced optimization framework for interpreting GCN and its variants such as APPNP, JKNet, DAGNN, and
GNN-LF/HF \cite{zhu2021interpreting}, by introducing appropriate regularizers.

Within these interpretable frameworks, some researchers have observed potential interference between node attributes and topology structures, especially in complex graph data. Node attributes are susceptible to disruption by the topology structure \cite{yang2019dual,yang2022graph}. GNNs tend to smooth the node attribute through topology, leading to the over-smoothing problem when multi-layer of GNNs are stacked.  The core issue with over-smoothing phenomenon is that topology excessively interferes with node attribute, diminishing the discriminability of embedding representations. In addition, in graph data with inaccurate topology, message propagation also enables nodes with similar attributes to obtain differential embedding representations \cite{jin2020graph}. Conversely, node attributes may also negatively interfere with topology. In link prediction tasks, topology information plays dominant role\cite{zhang2018link}. GNNs encoding both attribute and topology exhibit inferior performance compared to matrix factorization and node2vec \cite{grover2016node2vec} methods that solely utilize topology information. From an optimization perspective, GNNs learns a unified representation by 
compromising attribute and topology information. Due to the essential differences in attribute and topology, GNNs struggle to achieve a harmonious balance between them\cite{yang2019dual,wang2020gcn,yang2022graph}. Consequently, GNNs only explore partial and shared information among attribute and topology, resulting in the loss of complementary information.

Recently, to reduce the interference of topology on attribute, Dual Self-
Paced Graph Convolutional Network (DSP-GCN) \cite{yang2019dual} exploits the edge-level and node-level self-paced learning mechanisms. The DSP-GCN gradually introduces edges from simple to complex rather than the entire topology during the network training. With the optimization framework of GNNs, GNNs Beyond Compromise (GNN-BC) \cite{yang2022graph}  seeks to overcome existing compromises by utilizing various representations to severally preserve the attribute and topology information. Further, to reduce redundancy, GNN-BC compulsorily restricts representations of node attribute and topology to be exclusive by employing Hilbert-Schmidt Independence Criterion (HSIC) \cite{gretton2005measuring}. GNN-BC achieves excellent performance in some datasets, especially for heterogeneous graph. However, for homogeneous graph, its performance is mediocre. This may be primarily attributed to the HSIC constraint inducing GNN-BC to learn completely opposite representations, violating the principles of representation complementarity and structural consistency in multiple representation learning. More importantly, compared to the direct relationship at the feature level, representations from different perspectives have a stronger correlation at the structural level, which is more conducive to improving the flexibility of representations.

Therefore, this paper comes up with
a Decoupled Graph Neural Networks (DGNN) framework, which explores specific embedding representations for attribute and graph structure in a decoupled manner to counter adversarial interference among them. The overall framework is illustrated in Fig.~\ref{framework}. The DGNN explores the structural consistency between reconstructed adjacency matrices of various representations, and a shared latent reconstruction factor is utilized to promote semantic consistency. Thus, the DGNN, within the optimization framework, iteratively updates different representation to explore complementary information. A comprehensive and powerful representation for node-classification task is achieved by concatenating respective representations of attribute and graphs.

We summarize the main contributions of our paper from the following three aspects:
\begin{itemize}
    \item We propose a novel optimization framework induced DGNN to alleviate adversarial interference between attribute and graph during the embedding representation learning process. DGNN concurrently learns various embedding representations, breaking the compromise at the feature level.
    \item To promote representation complementarity and semantic consistency among different representations, we establish tight
connections  between  their reconstruction adjacency matrices using a  shared latent reconstruction factor. This structural correlation exploration reduces feature redundancy and improves the flexibility of representation learning.
   \item We employ an alternating optimization schema to iteratively update variables in DGNN. The extensive experiments conducted on real-world datasets sufficiently verify the effectiveness of DGNN in node classification task.
\end{itemize}

The rest of the paper is organized as follows. Section~\ref{Sec:3} reviews CNNs and their optimization framework. In Section~\ref{Sec:4}, we formally define our DGNN within a optimization framework and provide detailed explanation and its optimization algorithm. We report extensive experimental results in Section~\ref{Sec:5}.
Finally, Section~\ref{Sec:6} concludes the work and proposes the future direction.

\section{Preliminary}\label{Sec:3}

Before elaborating our approach in detail, we first introduce certain notations and classical graph neural networks. Additionally, we analyze the GNNs from an optimization framework perspective to highlight their limitation.

\subsection{Notations}
Let $\mathcal G=(\mathcal V, \mathcal E, \mathbf X)$ denote an undirected graph with a node set $\mathcal V=\{v_1, \cdots, v_N\}$, an edge set $\mathcal E$, and node attribute features $\mathbf X = [\mathbf x_1, \mathbf x_2, \cdots, \mathbf x_N]^T \in \mathbb R^{N \times D}$, where $N$ is the number of the nodes. The connection relationships of graph $\mathcal G$ can be fully described as the adjacency matrix $\mathbf A = [\mathbf A_{ij}]_{i,j=1}^N \in \{0,1\}^{N \times N}$, and each element in $\mathbf A$ represents the connectivity relationships between corresponding nodes, i.e.,
\begin{equation}
\mathbf A_{ij}=
\begin{cases}
1 & \text{if} \ \ (v_i,v_j) \in \mathcal E;\\
0 & \text{otherwise}.
\end{cases}
\end{equation}

The adjacency matrix $\mathbf A$ captures the physical topological structure of
nodes. In fact, the semantic connection relationships derived from attribute feature is a beneficial supplement to topological structure, improving the   flexibility of embedding representation learning.  To measure feature similarity among nodes, we construct the similarity matrix $\mathbf S_f = [{\mathbf S_f}_{ij}]_{i,j=1}^N \in \mathbb R^{N \times N}$ as follows:
\begin{equation}
    {\mathbf S_f}_{ij} = \frac{\langle \mathbf x_i, \mathbf x_j \rangle}{\|\mathbf x_i\| \cdot \|\mathbf x_j\|}.
\end{equation}
Then, we can construct the feature-based semantic graph adjacency matrix $\mathbf A_f$ by choosing the top-$K$ neighbors as the connected edges for a target node according to the similarity matrix $\mathbf S_f$.

\subsection{Graph Neural Networks}
The topological graph structure $\mathbf A$ reveals the intrinsic connection relationships
among nodes, providing  valuable guidances for representation learning across various tasks. GNNs have the capacity to learn discriminative and comprehensive node representations by encoding both attribute feature and graph structure. These networks aim to facilitate information propagation  through the aggregation and transformation of node feature. Various information propagation schemes exist, such as GCN, APPNP, and SGC.
Due to its strong ability of neighborhood information aggregation, GCN has attracted significant attention  across diverse applications and has demonstrated satisfactory performance.

The classical GCN encodes both topological graph information and attribute features to learn graph embedding representation by the following graph convolution operation:
\begin{equation}\label{gcn}
    f(\mathbf X, \mathbf A; \mathbf W) = \phi\left(\widetilde{\mathbf{D}}^{-\frac{1}{2}} 
    \widetilde{\mathbf A} \widetilde{\mathbf{D}}^{-\frac{1}{2}} \mathbf X \mathbf {W}\right),
\end{equation}
where $\tilde{\mathbf{A}}=\mathbf{A}+\mathbf{I}$ is the modified adjacency matrix with self-loops, $\widetilde{\mathbf{D}}$ is a diagonal degree  matrix, in which the diagonal elements of $\mathbf D$ represent the number of edges connected to the target node $v_i$, i.e., $\widetilde{\mathbf{D}}_{ii}=\sum\limits_{j=1}^n \tilde{\mathbf{A}}_{{ij}}$. In \eqref{gcn},  $\widetilde{\mathbf{D}}^{-\frac{1}{2}} \widetilde{\mathrm{\mathbf A}} \widetilde{\mathbf{D}}^{-\frac{1}{2}}$ is the normalized adjacency matrix, $\mathbf{W} \in \mathbb R^{D \times d}$ is the learnable parameters for feature transformation, and $\phi$ is a non-linear activation function to improve the representation ability of GCN. GCN can be explained as the neighborhood aggregation and message passing. The feature transformation and aggregation operation in \eqref{gcn} is as follows:
\begin{equation}
\text{Feature Transformation:} \ \mathbf{X'} =\mathbf{X}\mathbf{W};
\end{equation}
\begin{equation}
\text{Feature Aggregation:} \ \mathbf{H} =\widetilde{\mathbf{D}}^{-\frac{1}{2}} \widetilde{\mathbf A} \widetilde{\mathbf{D}}^{-\frac{1}{2}}\mathbf{X'}.
\end{equation}
Thus, GCN with multi-layer graph convolutions can capture higher-order structural information and learn more robust node embedding representation.

\subsection{Optimization Perspective Analysis of GNNs}

To understand various GNNs, researchers try to explain them from a   unified optimization perspective. Ma et al. \cite{ma2021unified} summarize several classical GNN models (such as GCN, GAT, and APPNP) as graph signal denoising optimization problems with different forms of Laplacian regularizer (including unnormalized and normalized versions).

\begin{Def}[\textbf{Graph Signal Denoising (GSD)}] Given a noisy signal $\mathbf S \in \mathbb R^{N \times D}$ on the graph $\mathcal G$, the goal of GSD is to learn a clean and smoother graph signal $\mathbf F \in \mathbb R^{N \times D}$, which, mathematically, is the solution to the following optimization problem % of graph signal denoising problem is formulated as follows:
\begin{equation}\label{gsd}
\min\limits_{\mathbf{F}} \mathcal{L} = \|\mathbf {F}-\mathbf {S}\|_{F}^{2} + \lambda \mathrm{tr}(\mathbf F^T \widetilde{\mathbf L} \mathbf F),
\end{equation}
where $\widetilde{\mathbf L}$ is the graph Laplacian matrix. % with different form on the graph $\mathcal G$.
\end{Def}
%\GaoC{Why with different form? Laplacian matrix is a specially defined matrix.}

In~\eqref{gsd}, the first objective item ensures that $\mathbf F$ should be close to the noisy signal $\mathbf S$. The second item is the graph Laplacian regularizer, which promotes a recovered smoother $\mathbf F$ in the sense with smaller Dirichlet energy. % lying on a smooth graph manifold, i.e., 
A smaller Dirichlet energy ensures that, if node $v_i$ and $v_j$ are connected on the graph $\mathcal G$, the corresponding signals $\mathbf F_{i,:}$ and $\mathbf F_{j,:}$ are similar. The objective function $\mathcal L$ is convex with respect to $\mathbf F$, and thus stationary point of optimization problem \eqref{gsd} can be solved by setting the first derivative to $\mathbf 0$, i.e.,
\begin{equation}
    \frac{\partial\mathcal{L}}{\partial\mathbf{F}} = 2(\mathbf F - \mathbf S)+ 2\lambda \widetilde{\mathbf L} \mathbf F = \mathbf 0,
\end{equation}
ans thus
\begin{equation}\label{upF}
    \mathbf F = (\mathbf I + \lambda \widetilde{\mathbf L})^{-1}\mathbf S.
\end{equation}
The solution~\eqref{upF} for $\mathbf F$ involves an matrix inverse operation on a matrix of size $N\times N$, which is $O(N^3)$. Here, if we set $\lambda=1$, $\mathbf S = \mathbf X'$, $\widetilde{\mathbf L} = \mathbf I-\widetilde{\mathbf{D}}^{-\frac{1}{2}} \widetilde{\mathbf A} \widetilde{\mathbf{D}}^{-\frac{1}{2}}$ and approximate the matrix inverse $(\mathbf I + \lambda \widetilde{\mathbf L})^{-1}$ by the first-order expansion, i.e., $(\mathbf I + \lambda \widetilde{\mathbf L})^{-1} \approx \mathbf I - \lambda \widetilde{\mathbf L}$, thus Eq.~\eqref{upF} would degrade into the updating rule of GCN, i.e.,
\begin{equation}\label{solution2}
    \mathbf F = \widetilde{\mathbf{D}}^{-\frac{1}{2}} \widetilde{\mathbf A}\widetilde{\mathbf{D}}^{-\frac{1}{2}}\mathbf{X'}.
\end{equation}

The approximated solution \eqref{solution2} of GSD problem is consistent with feature aggregation operator in GCN. According to the above analysis, we can conclude that GCN with
multiple graph convolution layers is equivalent to circularly solving the GSD problem with various noisy signals multiple times.

\section{The Proposed Model}\label{Sec:4}
In this section, we begin by scrutinizing the current limitations of GNNs from the optimization perspective. Subsequently, we develop
a novel optimization framework induced GNN, i.e. Decoupled GNN (DGNN), to address the existing issues. DGNN learns specific embedding representations for attribute
and graph in a decoupled manner. We also provide an effective optimization and implementation along with detailed complexity analyses.

\subsection{Motivation}
Building upon the optimization perspective analysis of Graph Neural Networks (GNNs presented above), the fundamental objective of GNNs lies in learning a consensus embedding representation that approximates the original attribute matrix $\mathbf X$ and preserves the intrinsic pairwise node similarity within the given graph.  This implies that the embedding representation serves as a delicate balance between attribute feature and graph structure. While such a balanced embedding representation captures both attribute information and graph structure, it proves advantageous for subsequent tasks. %This balanced embedding representation encodes both the attribute information and graph information, which appears to be beneficial for subsequent tasks. 
However, attribute feature and graph structure represent distinct characteristics of the original nodes, introducing inherent differences. The potential interference between attribute feature and graph structure inevitably weakens the expressiveness and discriminative power  of embedding representation. On one hand, the mandatory pursuit of a consensus embedding representation only explores partial useful information shared by attribute and graph structure, resulting in a deficiency in exploring complementary information.
On the other hand, the performance of GNNs is sensitive to the quality of graph structure. The actual graph structure is usually noisy due to carefully-crafted attacks. In addition, many existing graph are  heterogeneous,  meaning that  paired nodes with dissimilar attribute features are more likely to be connected than those with similar attributes. Obviously, GNNs with a focus on feature smoothness are not suitable for graph exhibiting high heterophily.

\subsection{The Framework of Decoupled Graph Neural Network}

To eliminate interference between node attributes and graph structure, we propose the Decoupled Graph Neural Networks (DGNN), which learns specific embedding representations for them and establishes a soft connection at a structural level rather than representation consistency. 

Inspired by GSD induced GNNs framework \eqref{gsd}, the output of the DGNN layer can be formulated the solution for the following optimization problem:

%the general framework of DGNN is formulated as follows:
\begin{equation}\label{graph3}
\min\limits_{\mathbf{F},\mathbf{H}} \mathcal{L} =\|\mathbf {F}-\mathbf {X}\|_{F}^{2} + \lambda \text{tr}(\mathbf {H}^T \widetilde{\mathbf{L}}\mathbf{H}) +\beta \Omega(\mathbf F, \mathbf H),
\end{equation}
where $\lambda$ and $\beta$ are non-negative parameter to balance the different objective items. In \eqref{graph3}, the DGNN learns different embedding representations to fit the node attribute and graph structure, respectively. Thus, the DGNN can avoid underlying interference between attribute features and graph structure and obtain complementary embedding representations. However, attribute feature and graph structure serve as crucial  descriptors of the original node from different perspectives. Exploring the correlation between $\mathbf F$ and $\mathbf H$ proves advantageous in minimizing redundancy and fostering the learning of discriminative embedding representations. With this goal in mind, $\Omega(\cdot)$ in \eqref{graph3} represents some feasible regularizers employed to establish the connection between the attribute embedding representation $\mathbf F$ and the graph embedding representation $\mathbf H$.

Some approaches  aim to uncover latent correlations at the feature level. For instance, if $\Omega(\mathbf F, \mathbf H) = \|\mathbf F - \mathbf H\|_F^2$ and $\beta$ is large enough, the optimal solution of \eqref{graph3}  leads to $\mathbf F = \mathbf H$, essentially causing DGNN to degenerate into the classical GCN with GSD optimization framework. Yang et al. \cite{yang2022graph} utilize the Hilbert-Schmidt Independence
Criterion as a mutual exclusion constraint to exploit ample information. 

In fact, when contrasted with rigid correlations among learned features, the structural correlation between embedding representations from different perspectives proves to be more robust and rational.  This insight motivates us to explore the representation complementarity and the underlying structural consistency between $\mathbf F$ and $\mathbf H$.

Therefore, we proceed to reconstruct the adjacency matrices from diverse embedding representations,  aiming to ensure structural consistency. The structural consistency regularizer for the adjacency reconstruction of embedding representations $\mathbf F$ and $\mathbf H$ is defined as 
\begin{equation}\label{sc}
    \Omega(\mathbf F, \mathbf H) = \|\mathbf F \mathbf F^T - \mathbf H \mathbf H^T\|_F^2,
\end{equation}
where $\mathbf F \mathbf F^T \in \mathbb R^{N \times N}$ and $\mathbf H \mathbf H^T \in \mathbb R^{N \times N}$ are the reconstructed adjacency matrices, which approximately utilize the cosine
similarity to measure the connections among nodes. Thus, the inclusion of the structural consistency term \eqref{sc} in DGNN enables the learning of complementary representations, facilitating a more effective exploration of connective relationships  between nodes  through distinct  representations $\mathbf F$ and $\mathbf H$, thereby achieving semantic information consistency. However, the connection information in the reconstructed adjacency matrix often contains redundancies, and the structural consistency constraint \eqref{sc} is too rigid. To refine the adjacency reconstruction, we propose the parameterized structural consistency constraint, defined as
\begin{equation}\label{sc1}
    \Omega(\mathbf F, \mathbf H) = \|\mathbf F \mathbf W_s \mathbf F^T - \mathbf H \mathbf W_s\mathbf H^T\|_F^2,
\end{equation}
where $\mathbf W_s \in \mathbf R^{D \times D}$ is the learnable and shared  reconstruction factor matrix. 

The shared reconstruction factor $\mathbf W_s$ plays a crucial role in  establishing
the underlying correlation between two reconstruction adjacency matrices, implicitly
exploring similar semantic information. However, the objective function in \eqref{sc1} still remains imprecise, because  it overlooks the symmetry of reconstructed adjacency
matrix. This oversight violates the property that the adjacency
matrix should reflects the connection relationships of
the target node from both the corresponding row and column in the adjacency
matrix. Thus, we further consciously the constrain $\mathbf W_s = \mathbf W \mathbf W^T$ to ensure symmetry. Consequently, the reformulated objective function in \eqref{graph3} becomes
\begin{equation}\label{dgnn1}
\min\limits_{\Upsilon} \|\mathbf {F}-\mathbf {X}\|_{F}^{2} + \lambda \text{tr}(\mathbf {H}^T \widetilde{\mathbf{L}}\mathbf{H}) +\beta \|\mathbf F \mathbf W_s \mathbf F^T - \mathbf H \mathbf W_s\mathbf H^T\|_F^2,
\end{equation}
where $\Upsilon = \{\mathbf F, \mathbf H, \mathbf W\}$ are optimization variables and $\mathbf W_s = \mathbf W \mathbf W^T$, $\mathbf W \in \mathbb R^{D \times D}$, is the learned factor. 

%Similar to GNNs with GSD optimization framework \eqref{gsd}, the formulation in 
We refer to \eqref{dgnn1} the optimization framework, from which our DGNN can be derived. DGNN is designed to circumvent interference between two embedding representations at the feature level and establish an underlying correlation at the structure level. The
structural consistency component facilitates mutual promotion and correction of these representations, improving the adaptability of DGNN across various tasks.

The DGNN formulation in \eqref{dgnn1} primarily focuses on exploring  topological structure, while the
semantic structure embedding in attribute feature is overlooked. This becomes crucial for achieving more versatile and comprehensive  embedding representation learning, particularly in scenarios where the topological graph is inaccurate or heterogeneous. Consequently, we extend our approach to learn dual graph embedding representations from both topological and semantic structures. Subsequently, we aim to unravel the complex structural consistency between one attribute embedding and two graph embeddings. The final objective function of a implicit layer in DGNN is as follows
\begin{equation}\label{dgnn}
\begin{aligned}
\min\limits_{\Upsilon} \ &\|\mathbf {F} -\mathbf {X}\|_{F}^{2}+\lambda \text{tr}(\mathbf H^T \widetilde{\mathbf L}\mathbf H) + \alpha \text{tr}(\mathbf H_f^T \widetilde{\mathbf L}_f\mathbf H_f) \\
&+ \beta \left\| \mathbf{F}\mathbf{W}_s\mathbf{F}^T -\left[\varepsilon\mathbf{H}\mathbf{W}_s\mathbf{H}^T +(1-\varepsilon)\mathbf{H}_f\mathbf{W}_s\mathbf{H}_f^T \right]\right\|_F^2
\end{aligned}
\end{equation}
where $\Upsilon = \{\mathbf{F}, \mathbf{H},\mathbf{H}_f, \mathbf W\}$ are optimization variable; the third term is feature-based semantic graph Laplacian regularizer to obtain embedding representation $\mathbf H_f$, $\widetilde{\mathbf L}_f = \mathbf I-\widetilde{\mathbf{D}}_f^{-\frac{1}{2}} \widetilde{\mathbf A}_f \widetilde{\mathbf{D}}_f^{-\frac{1}{2}}$, in which $\widetilde{\mathbf A}_f = \mathbf A_f+\mathbf I$, $\widetilde{\mathbf{D}}_f$ is degree matrix of $\widetilde{\mathbf A}_f$; $0 \leq \varepsilon \leq 1$ is a balance coefficient, and we empirically set it to $0.5$. 

Considering that the adjacency matrix
reconstructed by $\mathbf H_f$ contains complementary and valuable connection information, we combine the two adjacency matrices reconstructed by various graph embedding representation $\mathbf H$ and $\mathbf H_f$ together. This combined reconstruction approach provides  a comprehensive relationship, enabling the accurate establishment of  structural consistency between attribute and graph representations. The DGNN with multiple layers can be equivalently converted into solving optimization problem~\eqref{dgnn} multiple times, in which the output of the previous layer is set as the input for the subsequent layer. The detailed process  will be shown in Section~\ref{optmization}.

Once the optimization problem of DGNN is solved, we can obtain specific embedding representations $\{\mathbf F, \mathbf H, \mathbf H_f\}$ for original attribute feature $\mathbf X$, topological graph $\mathbf A$, and semantic graph $\mathbf A_f$, respectively. This approach mitigates conflict and accommodation between descriptions from different perspectives, resulting in complementary embedding representations. A more comprehensive and powerful representation for downstream tasks is obtained by concatenating these embeddings, i.e.,
\begin{equation}
    \mathbf Z = \left[\mathbf F, \mathbf H, \mathbf H_f\right].
\end{equation}
The process of separate learning and concatenation helps the classifier distinguish the importance of different embedding representations, thereby improving the flexibility and discrimination of classifier. 

To pipeline the concatenated features $\mathbf Z$ to a classifier, we define a fully connected layer as follows, %Mathematically, the feature transformation of the fully connected layer in the classifier is formulated as
\begin{equation}\label{classifier}
\begin{aligned}
    \mathbf Z \mathbf W_c = &[\mathbf F, \mathbf H, \mathbf H_f][\mathbf W_1, \mathbf W_2, \mathbf W_3]^T \\
    = & \mathbf F \mathbf W_1^T + \mathbf H \mathbf W_2^T + \mathbf H_f \mathbf W_3^T,
\end{aligned}
\end{equation}
where $\mathbf W_c = [\mathbf W_1, \mathbf W_2, \mathbf W_3]^T \in \mathbb R^{3D \times c}$ is learnable parameter matrix in the fully connected layer. If the $j$-th column of the parameter matrix $\mathbf W_i$ $(i=1, 2, 3)$ is zero, it means that the corresponding feature in embedding representations $\{\mathbf F, \mathbf H, \mathbf H_f\}$ does not contribute to the classifier. Thus, the parameter matrices $\{\mathbf W_1, \mathbf W_2, \mathbf W_3\}$ can be regarded as the importance metric weights, and our DGN could adaptively distinguish $\mathbf F$, $\mathbf H$, and $\mathbf H_f$ and achieve feature selection by back propagation during the training process.

For semi-supervised node classification task, the predicted soft label is obtained as follows
\begin{equation}
    \mathbf Y^{pre}=\text{softmax}(\mathbf Z \mathbf W_c+ \mathbf b) \in \mathbb R^{N \times c},
\end{equation}
where $\mathbf Y_{ik}^{pre}$ is the probability that node $v_i$ belongs to the $k$-th class. Thus, the loss function of training is defined as follows
\begin{equation}
    \mathcal L_c = \sum\limits_{v_i \in \mathcal V_L} l(\mathbf Y_{i,:}^{pre},\mathbf Y_i),
\end{equation}
where $\mathbf Y_i$ is the ground truth of node $v_i$, $\mathcal V_L$ is the labeled training set, and $l(\cdot)$ is the cross entropy function.

\subsection{Optimization}\label{optmization}
The alternating iteration algorithm is utilized to solve the
optimization problem \eqref{dgnn} with multi-variables. Specifically, we alternatively
optimize each variable while the others are fixed. For clarity, the objective function of \eqref{dgnn} is noted as $\mathcal L(\mathbf F, \mathbf H, \mathbf H_f)$.

\textbf{Update} $\mathbf F$: Fixing the other variables, take the derivative of $\mathcal L(\mathbf F, \mathbf H, \mathbf H_f)$ w.r.t. $\mathbf F$ and set it to zero, i.e.,
\begin{equation}\label{updateF}
\begin{aligned}
 &\frac{\partial\mathcal{L}(\mathbf F, \mathbf H, \mathbf H_f)}{\partial\mathbf{F}} \\
= 
& \ 2(\mathbf{F}-\mathbf{X})+2\beta \left[\mathbf{F}\mathbf{W}_s\mathbf{F}^T -\varepsilon\mathbf {H}\mathbf W_s\mathbf {H}^T \right.\\
&\left. -(1-\varepsilon)\mathbf {H}_f \mathbf {W}_s \mathbf {H}_f^T \right]\left(\mathbf{F}\mathbf{W}_s + \mathbf{F}\mathbf{W}_s^T \right)\\
= & \ \mathbf 0.
\end{aligned}
\end{equation}
Thus, we can solve $\mathbf F$ by the following iterative rule:
\begin{equation}
\begin{aligned}
    \mathbf F = \ &\mathbf X - \beta \left[\mathbf{F}\mathbf{W}_s\mathbf{F}^T -\varepsilon\mathbf {H}\mathbf W_s\mathbf {H}^T \right.\\
&\left. -(1-\varepsilon)\mathbf {H}_f \mathbf {W}_s \mathbf {H}_f^T \right]\left(\mathbf{F}\mathbf{W}_s + \mathbf{F}\mathbf{W}_s^T \right).\\
\end{aligned}
\end{equation}

\textbf{Update} $\mathbf H$: Fixing the other variables, take the derivative of $\mathcal L(\mathbf F, \mathbf H, \mathbf H_f)$ w.r.t. $\mathbf H$ and set it to zero, i.e.,
\begin{equation}\label{updateH}
\begin{aligned}
 &\frac{\partial\mathcal{L}(\mathbf F, \mathbf H, \mathbf H_f)}{\partial\mathbf{H}} \\
= 
& \ 2\lambda \widetilde{\mathbf L}\mathbf H+2\varepsilon \beta \left[\varepsilon\mathbf {H}\mathbf W_s\mathbf {H}^T+(1-\epsilon)\mathbf {H}_f \mathbf {W}_s \mathbf {H}_f^T  \right.\\
&\left. -\mathbf{F}\mathbf{W}_s\mathbf{F}^T \right]\left(\mathbf{H}\mathbf{W}_s + \mathbf{H}\mathbf{W}_s^T \right)\\
= & \ \mathbf 0.
\end{aligned}
\end{equation}
Thus, we can solve $\mathbf H$ by the following iterative rule:
\begin{equation}
\begin{aligned}
    \mathbf H =\ & \hat{\mathbf A}\mathbf H -\varepsilon \frac{\beta}{\lambda} \left [ \varepsilon \mathbf H \mathbf W_s \mathbf H^T+(1-\varepsilon)\mathbf H_f \mathbf W_s \mathbf H_f^T\right. \\
    & \left. -\mathbf F \mathbf W_s \mathbf F^T \right]\left(\mathbf{H}\mathbf{W}_s + \mathbf{H}\mathbf{W}_s^T\right),
\end{aligned}
\end{equation}
where $\hat{\mathbf A} = \widetilde{\mathbf{D}}^{-\frac{1}{2}} \widetilde{\mathbf A} \widetilde{\mathbf{D}}^{-\frac{1}{2}}$.

\textbf{Update} $\mathbf H_f$: Fixing the other variables, take the derivative of $\mathcal L(\mathbf F, \mathbf H, \mathbf H_f)$ w.r.t. $\mathbf H_f$ and set it to zero, i.e.,
\begin{equation}\label{updateH_f}
\begin{aligned}
 &\frac{\partial\mathcal{L}(\mathbf F, \mathbf H, \mathbf H_f)}{\partial\mathbf{H}_f} \\
= 
& \ 2\alpha \widetilde{\mathbf L}_f\mathbf H_f+2(1-\varepsilon) \beta \left[\varepsilon\mathbf {H}\mathbf W_s\mathbf {H}^T+(1-\varepsilon)\mathbf {H}_f \mathbf {W}_s \mathbf {H}_f^T  \right.\\
&\left. -\mathbf{F}\mathbf{W}_s\mathbf{F}^T \right]\left(\mathbf{H}_f\mathbf{W}_s + \mathbf{H}_f\mathbf{W}_s^T \right)\\
= & \ \mathbf 0.
\end{aligned}
\end{equation}
Thus, we can solve $\mathbf H_f$ by the following iterative rule:
\begin{equation}
\begin{aligned}
    \mathbf H_f = \ & \hat{\mathbf A}_f\mathbf H_f -(1-\varepsilon) \frac{\beta}{\alpha} \left [ \varepsilon \mathbf H \mathbf W_s \mathbf H^T+(1-\varepsilon)\mathbf H_f \mathbf W_s \mathbf H_f^T\right. \\
    & \left. -\mathbf F \mathbf W_s \mathbf F^T \right]\left(\mathbf{H}_f\mathbf{W}_s + \mathbf{H}_f\mathbf{W}_s^T\right),
\end{aligned}
\end{equation}
where $\hat{\mathbf A}_f = \widetilde{\mathbf{D}}_f^{-\frac{1}{2}} \widetilde{\mathbf A}_f \widetilde{\mathbf{D}}_f^{-\frac{1}{2}}$.

Instead of simply utilizing the explicit convolution operator~\eqref{gcn} as the network layer output, we impose the optimization problem~\eqref{dgnn} to implicitly define the new layer of DGNN. Further, an implicit layer can be unfolded by the iteration algorithm of the optimization problem~\eqref{dgnn} and the output of layer is its solution. Thus, the DGNN with $L$
layers is equivalent to iteratively solving problem~\eqref{dgnn} $L$ steps. To improve the non-linear representation ability and computational efficiency, we further introduce the activation function. Thus, for the $(k+1)$-th iteration, we start with $\mathbf F^{(k)}$, $\mathbf H^{(k)}$, $\mathbf H_f^{(k)}$ and
then alternately update the variables as follows:
\begin{equation}\label{updatef1}
\begin{aligned}
    &\mathbf F^{(k+1)}\\
    =\ & \mathbf X - \beta \sigma\left[\mathbf{F}^{(k)}\mathbf{W}_s{\mathbf{F}^{(k)}}^T -\varepsilon\mathbf {H}^{(k)}\mathbf W_s{\mathbf {H}^{(k)}}^T \right.\\
&\left. -(1-\varepsilon)\mathbf {H}_f^{(k)} \mathbf {W}_s {\mathbf {H}_f^{(k)}}^T \right]\left(\mathbf{F}^{(k)}\mathbf{W}_s + \mathbf{F}^{(k)}\mathbf{W}_s^T \right),\\
\end{aligned}
\end{equation}
\begin{equation}\label{updateh1}
\begin{aligned}
    &\mathbf H^{(k+1)} \\
    = \ & \hat{\mathbf A}\mathbf H^{(k)} -\varepsilon \frac{\beta}{\lambda}\sigma\left [ \varepsilon \mathbf H^{(k)} \mathbf W_s {\mathbf H^{(k)}}^T+(1-\varepsilon)\mathbf H_f^{(k)} \mathbf W_s {\mathbf H_f^{(k)}}^T\right. \\
    & \left. -\mathbf F^{(k)} \mathbf W_s {\mathbf F^{(k)}}^T \right]\left(\mathbf{H}^{(k)}\mathbf{W}_s + \mathbf{H}^{(k)}\mathbf{W}_s^T\right),
\end{aligned}
\end{equation}
\begin{equation}\label{updatehf1}
\begin{aligned}
    &\mathbf H_f^{(k+1)} \\
    = \ & \hat{\mathbf A}_f\mathbf H_f^{(k)} -(1-\varepsilon) \frac{\beta}{\alpha} \sigma\left [-\mathbf F^{(k)} \mathbf W_s {\mathbf F^{(k)}}^T + \varepsilon \mathbf H^{(k)} \mathbf W_s {\mathbf H^{(k)}}^T\right. \\
    & \left. +(1-\varepsilon)\mathbf H_f^{(k)} \mathbf W_s {\mathbf H_f^{(k)}}^T\right]\left(\mathbf{H}_f^{(k)}\mathbf{W}_s + \mathbf{H}_f^{(k)}\mathbf{W}_s^T\right),
\end{aligned}
\end{equation}
where $\sigma(\cdot)$ is the sigmoid function for nonlinear transformation. After $L$ iterations, the attribute, topological graph, and semantic graph embedding representations $\mathbf F^{(L)}$, $\mathbf H^{(L)}$, and $\mathbf H_f^{(L)}$ are concatenated to construct a final and comprehensive representation for semi-supervised node classification task. And, the reconstruction factor $\mathbf W_s$ and classification parameters $(\mathbf W_c, \mathbf b)$ are updated by training the proposed DGNN with loss function $\mathcal L_c$.

\subsection{Complexity Analysis}
The overall computational complexity is mainly dominated by the updating rules \eqref{updatef1}, \eqref{updateh1}, and \eqref{updatehf1}. All updating rules only involve matrix addition and matrix-vector  multiplication. Thus, the complexity of updating rules is all $\mathcal O(DN^2+D^2N)$, which is quadratic with respect to the number of nodes. Hence, the computational complexity of the proposed DGNNs is comparable to widely used GCN methods. However, our DGNN could achieve more superior performance.

\section{Experiment and Analysis}\label{Sec:5}

In this section, we evaluate the performance of proposed DGNN method on thenode classification task. Our experiments are conducted on a Linux server with GPU (GeForce RTX A6000) and CPU (Intel(R) Core(TM) i9-10900X CPU @ 3.70 GHz). The code of our DGNN is written by PyTorch 1.7.1. The implementation and code of our DGNN would be available at \url{https://github.com/JinluWang1002/DGNN}.

\subsection{Experimental Datasets}
We conducted experiments to evaluate the effectiveness of our model using six widely used datasets. These datasets include two citation network datasets (\texttt{Cora} and \texttt{Citeseer}), two web page network datasets (\texttt{Chameleon} and \texttt{Squirrel}), and two co-purchase network datasets (\texttt{Photo} and \texttt{Computers}).

\textit{\textbf{Citation Networks:}} The citation networks are the most common graph data, in which nodes usually represent different papers and edges represent citation relationships between them. The node attribute is the bag-of-word feature of words in corresponding paper and node label is set according to the research topic of paper. The \texttt{Cora} consists of a total of 2708 papers, which are divided into 7 categories. The attribute of each paper is a 1433-dimensional word vector feature, and there are 5429 connected edges between all nodes. The \texttt{Citeseer} is another classic citation networks dataset, which includes 3327 scientific publications belonging to 6 categories. The dimension of the attribute features is 3703, and the citation network consists of 4732 links.

\textit{\textbf{Web Page Networks:}} These are page-page networks datasets from Wikipedia. The node in this graph networks denotes the article page from the English Wikipedia, the edges represents the mutual
links between web pages of the corresponding topics. The \texttt{Squirrel} is collected from Wikipedia,  which consists of 5201 nodes of 5 categories. The dimension of the attribute features is 2089, and there are 217073 connection relationships. The \texttt{Chameleon} is another web page networks datasets, including 2277 nodes and dividing into 5 categories. The dimension of the features is 2325, and it has 36101 connected edges.

\textit{\textbf{Co-purchase Networks:}} These are Amazon co-purchase relationship networks datasets. In these
networks, goods are represented as nodes of graph, and edges are
connected if two goods be frequently purchased together. The node attributes are product reviews with form of bag-of-words, and the product category is naturally set as the its label. The \texttt{Computers} dataset contains 13752 goods from 10 categories of computers. The dimension of the features is 767, and there are 245861 pairs of products being purchased simultaneously. The \texttt{Photo} dataset contains 7650 nodes divided into 8 categories. The dimension of the features is 745, and it has 119081 connection relationships.

The statistics of these benchmark
datasets are summarized in the TABLE~\ref{datasets}.

\begin{table}
    \renewcommand\tabcolsep{4pt}
  \centering
  \caption{The statistics of the used datasets.}\label{datasets}
    \begin{tabular}{cccccc}
\noalign{\smallskip}\hline\noalign{\smallskip}
    \textbf{Datasets} & \textbf{Nodes} & \textbf{Features} & \textbf{Classes} & \textbf{Edges} & \textbf{Homphily Rate}  \\
    \noalign{\smallskip}\hline\noalign{\smallskip}
    \textbf{Cora} & 2708 & 1433  & 7  & 5429 & 0.809\\
    \textbf{Citeseer} & 3327 & 3703 & 6  & 4732  & 0.721\\ 
    \textbf{Chameleon} & 2277 & 2325  & 5  & 36101  & 0.233 \\
     \textbf{Squirrel} & 5201 & 2089  &  5  & 217073  &  0.203\\
    \textbf{Computers} & 13752 & 767 & 10 &  245861 &  0.791 \\
    \textbf{Photo} & 7650  & 745 & 8   & 119081 & 0.824 \\
    \noalign{\smallskip}\hline\noalign{\smallskip}
    \end{tabular}
\end{table}

\subsection{Experimental Setting}

For all used datasets, following \cite{yang2022graph} and \cite{song2023ordered}, we randomly select $60\%$ labeled nodes as training set, $20\%$ nodes as validation set, and the remaining $20\%$ nodes as testing set.
There are three hyper-parameters $\lambda$, $\alpha$, and $\beta$ in our DGNN model, which are tuned according to accuracy on validation
set. The settings of these hyper-parameters, model layer $L$, dropout rate $dr$, and learning rate $lr$ on all datasets are shown in TABLE~\ref{hyperparameter}. We will demonstrate the impact of hyper-parameters $\lambda$, $\alpha$, and $\beta$ on the performance in node classification in Section~\ref{parameter_analysis}.

\begin{table}
   \renewcommand\tabcolsep{10pt}
  \centering
  \caption{The parameter setting of DGNN on all used datasets.}\label{hyperparameter}
    \begin{tabular}{ccccccc}
\noalign{\smallskip}\hline\noalign{\smallskip}
    \textbf{Datasets} & \textbf{$\lambda$} & \textbf{$\alpha$} & \textbf{$\beta$} & \textbf{$L$} & \textbf{ $dr$} & \textbf{ $lr$} \\
    \noalign{\smallskip}\hline\noalign{\smallskip}
    \textbf{Cora} & 1.0 & 2.0  & 0.02 & 2 & 0.25 & 0.002 \\
    \textbf{Citeseer} & 2.0 & 0.5 & 0.01  & 3  & 0.15 & 0.003 \\
    \textbf{Chameleon} & 1.0 & 2.5 & 0.01  & 2 & 0.02 & 0.05 \\
    \textbf{Squirrel} & 1.0 & 2.5  & 0.01  & 2  & 0.0 & 0.02 \\
    \textbf{Computers} & 2.0 & 1.0 & 0.01 & 2 & 0.05 & 0.03 \\
    \textbf{Photo} & 2.0 & 0.5 & 0.01 & 2  & 0.15 & 0.02\\
    \noalign{\smallskip}\hline\noalign{\smallskip}
    \end{tabular}
\end{table}

\subsection{Baselines}
To comprehensively verify the superiority of our DGNN, we compare it with 13 representative methods associated with GNNs on the node classification task. In our experiment, we select three 
%We can roughly categorize them into three categories, as follows:
\textbf{classical and fundamental GNNs models} as a benchmark, including  GCN\cite{kipf2017semi}, GAT \cite{velickovic2017graph}, and GraphSAGE \cite{hamilton2017inductive}. In addition, some \textbf{enhanced GNNs approaches} with state-of-art performance are selected as competitors, including GCNII \cite{chen2020simple}, APPNP \cite{gasteiger2018predict}, JKNet \cite{xu2018representation}, Geom-GCN \cite{pei2020geom}, GPRGNN \cite{chien2020adaptive}, FAGCN \cite{bo2021beyond}, H2GCN \cite{zhang2022h2gnn}, GNNBC \cite{yang2022graph}, and Ordered GNN \cite{song2023ordered}. The GCNII, APPNP, and JKNet can effectively alleviate over-smoothing problem in deep GNNs. The  Geom-GCN, GPRGNN, FAGCN, H2GCN, GNNBC, and Ordered GNN are proposed to be applicable to heterophilic graph data. To achieve fair comparisons, some original experimental results of compared methods are directly presented according to \cite{yang2022graph} and \cite{song2023ordered} due to the same experimental splitting.

\begin{table*}[t]
   \centering
    \renewcommand\arraystretch{1.7}
    \renewcommand\tabcolsep{12pt}
    \caption{The node classification results on the all used datasets. Remark: notation NA means that corresponding results are not reported in original manusript. }\label{result1}
    %\vspace{0.4cm}
   \begin{tabular}{c c c c c c c}
     \hline
      \textbf{Dataset} & \textbf{Cora} & \textbf{Citeseer} & \textbf{Chameleon} &  \textbf{Squirrel} & \textbf{Computers} & \textbf{Photo}\\
         \hline
              GCN & $85.77\pm0.25$ & $73.68\pm0.31$ & $28.18\pm0.23$ & $23.96\pm0.26$ & $82.52\pm0.32$ &  $90.54\pm0.21$\\
              GAT & $86.37\pm0.30$ & $74.32\pm0.27$ & $42.93\pm0.28$ & $30.03\pm0.25$ & $81.95\pm0.38$ &  $90.09\pm0.27$ \\
              GraphSAGE & $87.77\pm1.04$ & $71.09\pm1.30$ & $49.24\pm1.68$  & $36.28\pm1.73$ & $83.11\pm0.23$   & $90.51\pm0.25$ \\
              MLP & $74.82\pm2.22$ & $70.94\pm0.39$ & $49.67\pm0.78$ & $37.04\pm0.46$ & $70.48\pm0.28$ & $78.69\pm0.30$ \\
               \hline
              GCNII & $88.49\pm2.78$ & $77.08\pm1.21$ & $60.61\pm2.00$ & $37.85\pm2.76$ & $86.13\pm0.51$ & $90.98\pm0.93$ \\
              APPNP & $87.87\pm0.85$ & $76.53\pm1.33$ & $54.30\pm0.34$ & $33.29\pm1.72$ & $81.99\pm0.26$ & $91.11\pm0.26$\\
              JKNet & \textcolor{blue}{$88.93\pm1.35$} &	$74.37\pm1.53$ & $62.31\pm2.76$ & $44.24\pm2.11$ & $77.80\pm0.97$ &	$87.70\pm0.70$ \\
               \hline
              Geom-GCN-I & $85.19\pm1.13$ & \textcolor{blue}{$77.99\pm1.23$} & $60.31\pm1.77$ & $33.32\pm1.59$	& NA & NA \\
              Geom-GCN-P & $84.93\pm0.51$ & $75.14\pm1.50$ & $60.90\pm1.13$ & $38.14\pm1.23$ & NA	& NA  \\
              Geom-GCN-S & $85.27\pm1.48$	& $74.71\pm1.17$ & $59.96\pm2.03$ & $36.24\pm1.05$	& NA & NA  \\  
              GPRGNN & $88.65\pm1.37$ & $77.99\pm1.64$ & $67.48\pm1.98$ & $49.93\pm1.34$ & $82.90\pm0.37$ &  $91.93\pm0.26$ \\
              FAGCN  & $87.77\pm1.69$	& $74.66\pm2.27$ & $61.12\pm1.95$ & $40.88\pm2.02$ & $86.09\pm0.40$ & $91.96\pm0.71$ \\
              H2GCN-1 & $86.92\pm1.37$	& $77.07\pm1.64$ &  $57.11\pm1.58$ &  $36.42\pm1.89$ &	OOM & OOM \\
              H2GCN-2 & $87.81\pm1.35$ & $76.88\pm1.77$ &  $59.39\pm1.98$ & $37.90\pm2.02$ &	OOM	& OOM \\
              \hline
              GNNBC  & $88.75\pm1.21$ & $76.70\pm0.77$	& \textcolor{blue}{$74.63\pm0.93$}	& $61.41\pm1.55$	& \textcolor{blue}{$89.60\pm0.89$} & \textcolor{blue}{$93.17\pm0.67$} \\
              Ordered GNN & $88.37\pm0.75$ & $77.31\pm1.73$	& $72.28\pm2.29$ & \textcolor{blue}{$62.44\pm1.96$} & NA & NA \\
               \hline
              DGNN & \color{red}{$\mathbf{91.06\pm0.36}$} &  \color{red}{$\mathbf{78.39\pm0.30}$} &  \color{red}{$\mathbf{78.21\pm1.31}$} & \color{red}{$\mathbf{70.35\pm2.49}$} &  \color{red}{$\mathbf{91.70\pm0.18}$} & \color{red}{$\mathbf{93.98\pm0.16}$} \\
              \hline
   \end{tabular}\\[1.5mm]
\end{table*}

\subsection{Experimental Results and Analysis}

The node classification results of the proposed DGNN and all compared methods are presented in TABLE~\ref{result1}. The red bold \textcolor{red}{\textbf{scores}} indicate the best results and the blue \textcolor{blue}{\textbf{scores}} indicates the second best results. From the experimental results presented in TABLE~\ref{result1}, some interesting observations are summarized as follows.

From a global perspective, our DGNN method is consistently superior to all compared methods  across all used datasets. The performance improvement of DGNN is considerably significant on some datasets. For example, DGNN  achieves improvements of around $2.13\%$, $3.58\%$, $7.91\%$, $2.10\%$ over the corresponding second-best method for \texttt{Cora}, \texttt{Chameleon}, \texttt{Squirrel}, \texttt{Computers} datasets, respectively. These experimental observations thoroughly confirm the effectiveness of the proposed DGNN on the node classification task. And, across various application scenarios, including citation network, web page network, co-purchase network, etc, our DGNN invariably consistently attains optimal performance, validating its robustness for complex node classification applications.

Next, we analyze experimental results from a fine-grained level to fully demonstrate the rationality and superiority of the proposed DGNN. 
\begin{itemize}
    \item While GCNI, APPNP, and JKNet are designed to alleviate the over-smoothing problem and act as deep models, generally  surpassing classic GNNs methods (GCN, GAT, GraphSAGE),  our DGNN demonstrates clear superiority with substantial improvements. For example, DGNN achieves improvements of $17.60\%$, $32.50\%$, $5.57\%$ compared to GCNI on the \texttt{Chameleon}, \texttt{Squirrel}, \texttt{Computers} datasets, which verifies the advantages of DGNN in alleviating over-smoothing. This is mainly attributed to DGNN's ability to learn specific representations for both attribute and graph structures,  thereby explicitly enhancing node information through the concatenation of attribute and graph representations.  
    \item The node classification task on the \texttt{Chameleon} and \texttt{Squirrel} datasets is particularly challenging because their graph structure is high heterophily. Fortunately, our DGNN still achieves the best performance, notably outperforming Geom-GCN, GPRGNN, FAGCN, H2GCN, and Ordered GNN with evident improvements. Specifically, compared to the relatively stable methods GPRGNN and Ordered GNN, DGNN exhibits improvements of $10.73\%$ and  $5.93\%$ on the \texttt{Chameleon} dataset, $20.42\%$ and $7.91\%$ on the \texttt{Squirrel} dataset. This indicates the capability of DGNN to break compromises between node attribute and graph structure is particularly effective in mitigating   the negative influence of inaccurate graph structure in heterogeneous graph. In detail, as shown in \eqref{classifier}, DGNN automatically adjusts the contribution of different representations to the classifier, thereby reducing the impact of noisy representation.
   \item In addition, GNNBC serves as a direct and crucial benchmark, encoding the node attribute and topology with different embedding representations to alleviate the interference between them. Despite these efforts,  its performance still lags behind our DGNN on all datasets. GNNBC exhibits mediocre performance compared to other classic GNNs methods on homogeneous \texttt{Cora} and \texttt{Citeseer} datasets. The main reason  lies in GNNBC's approach of establishing relationships among embedding representations of attribute and topology from the feature level,  forcibly constraining them to be independent and resulting in mutually exclusive representations. In fact, node attribute and graph structure are both descriptors of original data from different views. The representation complementarity and structural consistency among their embedding representations are usually stronger. Fortunately, our DGNN explores the structural correlation between various embedding representations by unifying their adjacency reconstruction matrices with a shared reconstruction factor. The structural consistency constraint promotes the exploration of complementary information and semantic correlation among them,   effectively eliminating redundant information. Furthermore, feature-based semantic graph structure is utilized as topology refinement and complement, resulting in more comprehensive and accurate representations.

\end{itemize}

Therefore, the experimental results and detailed analyses evidently demonstrate the effectiveness and robustness of our DGNN for the node classification task.

\begin{figure*}
\centering
%\hspace{-10mm}
\subfigure [The results on \texttt{Cora}, \texttt{Computers}, and \texttt{Photo} datasets.]{
\includegraphics[width=8cm,height=6cm]{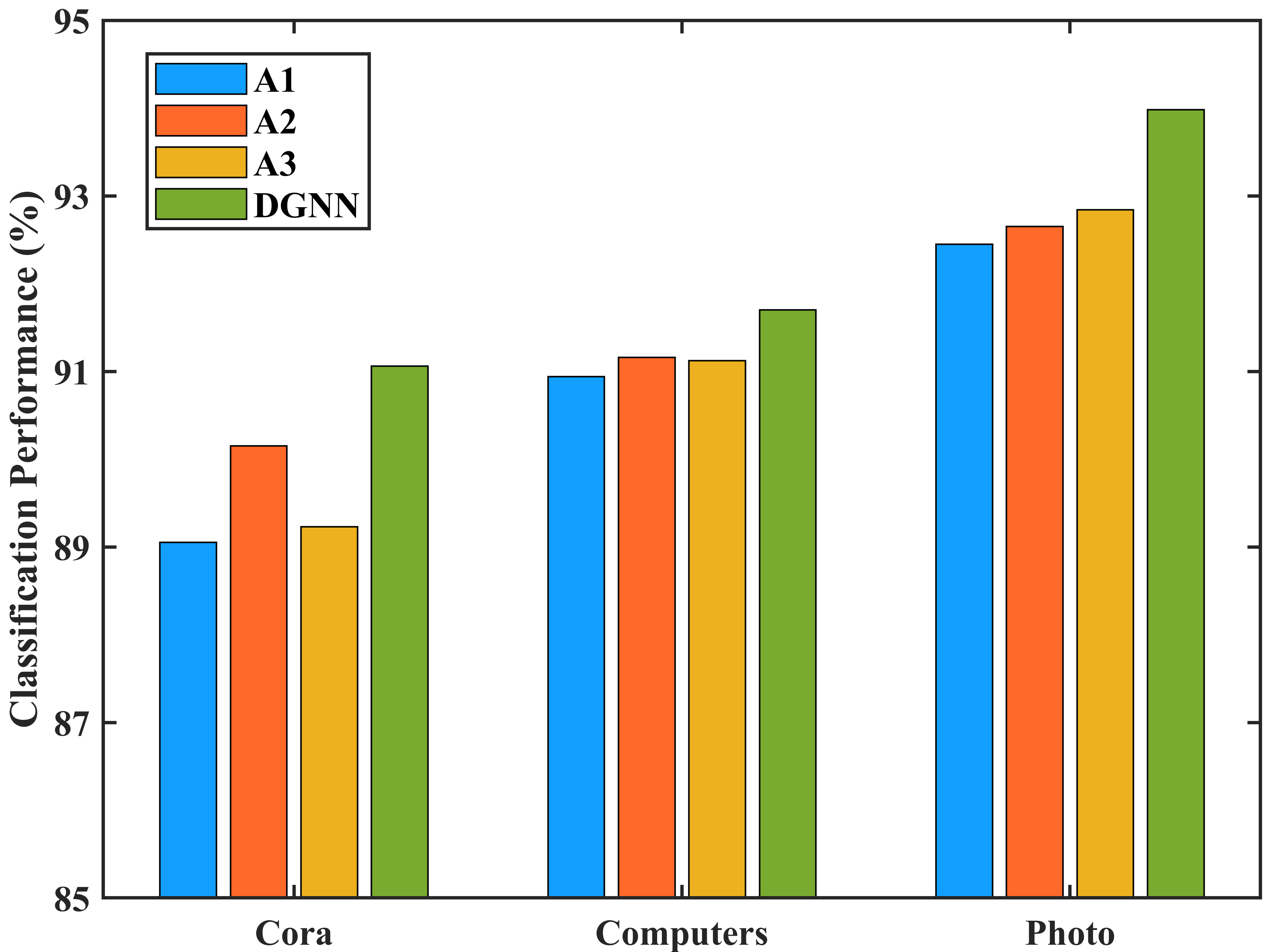}
}
\hspace{2mm}
\subfigure [The Results on \texttt{Citeseer}, \texttt{Chameleon}, and \texttt{Squireel} datasets.]{
\includegraphics[width=8cm,height=5.9cm]{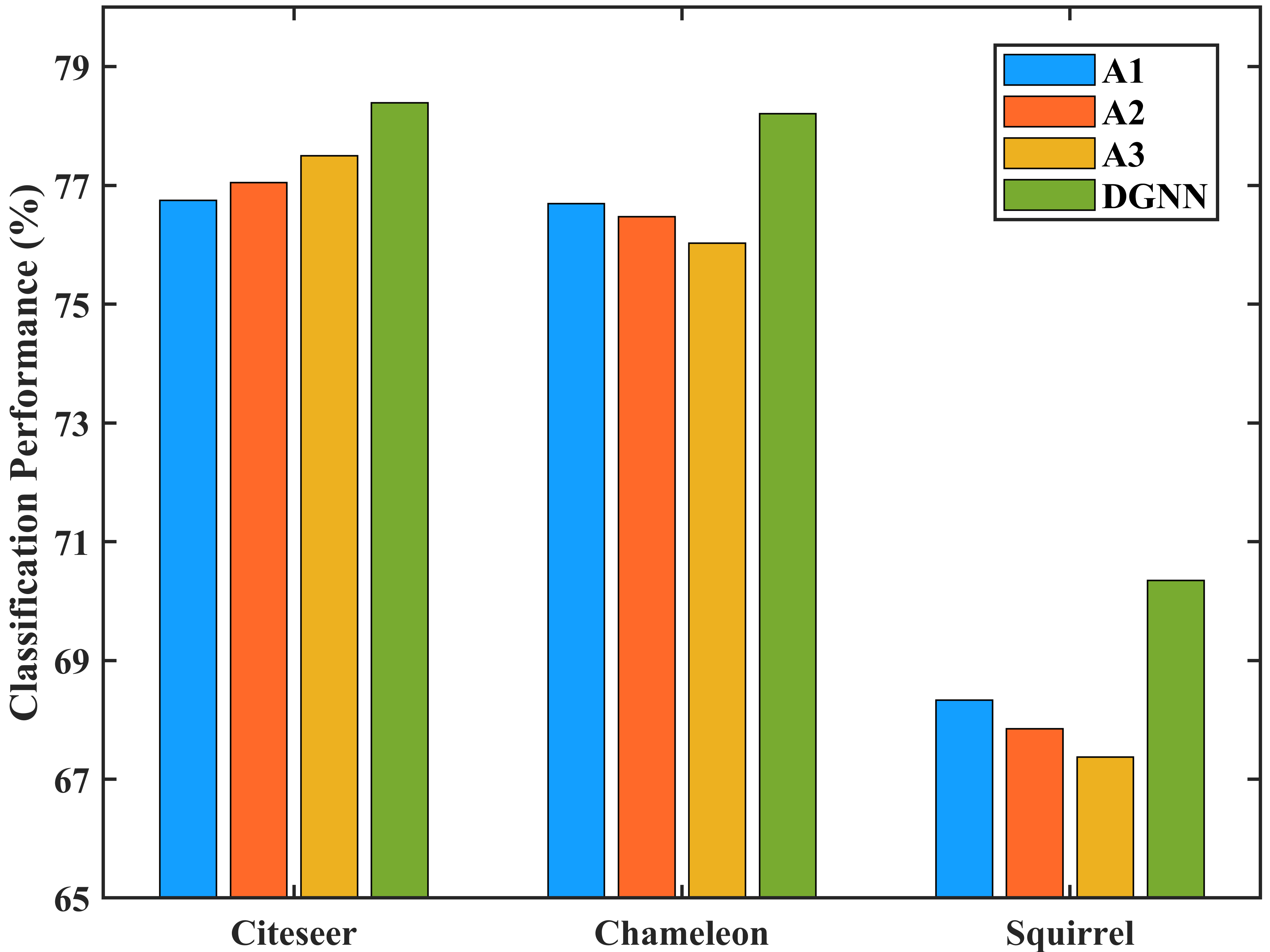}
}
\caption{The experimental results of all ablation model and DGNN model on the all datasets, where A1: DGNN with $\lambda=0$ and $\varepsilon=0$, A2: DGNN with $\alpha=0$ and $\varepsilon = 1$, A3: DGNN with $\beta=0$.}\label{ablation}
\end{figure*}

\subsection{Ablation Study}

To intuitively understand DGNN and validate the effectiveness of its components, we conduct an ablation study from the following aspects:
\begin{itemize}
    \item A1: whether the original topological graph is necessary and effective for generating powerful representations; 
     \item A2: whether the feature-based semantic graph is necessary for generating powerful representations, especially for heterogeneous graph; 
     \item A3: whether the structural consistency is beneficial for modeling flexible and accurate correlations among various embedding representations.
\end{itemize}

Specifically, DGNN with $\lambda=0$ and $\varepsilon=0$ is degenerated into ablation model A1; DGNN with $\alpha=0$ and $\varepsilon = 1$ is degenerated into ablation model A2; and DGNN with $\beta=0$ is degenerated into ablation model A3. The ablative experiment results are shown in Fig.~\ref{ablation}, leading to the following conclusions:
\begin{itemize}
    \item Overall, the complete DGNN method always outperforms  all the ablation models of DGNN, verifying the effectiveness of DGNN
and the necessity of each component for node classification tasks.
   \item For datasets with higher homogeneity such as  \texttt{Cora}, \texttt{Computers}, \texttt{Photo}, and \texttt{Citeseer}, the performance of A1 is slightly higher than that of A2. This suggests that the original topological graph can provide accurate and sufficient adjacency relationships. However, for  heterogeneous datasets like  \texttt{Chameleon} and \texttt{Squirrel}, the model A1 performs better than A2, directly verifying the necessity of feature-based semantic graph for discriminative representation learning. The performance of DGNN is inadequate when only topological graph or semantic graph is individually used. However, DGNN fully utilizes complementary information between them to better encode underlying structures and learn a comprehensive embedding representation.
   \item A3 performs even worse than the single-graph ablation model (A1 or A2) on some datasets, implying that reasonable modeling correlation among various representations crucial for effectively utilizing complementary information and removing redundancy. In our DGNN, the structural consistency term, with a shared latent reconstruction factor, is designed to  capture high-level correlations, ensuring the extraction of more flexible and discriminative embedding representations.
\end{itemize}

\subsection{Result Visualization}
\begin{figure*}
\centering
\subfigure [\texttt{Cora}-RAW]{
\includegraphics[width=4.2cm]{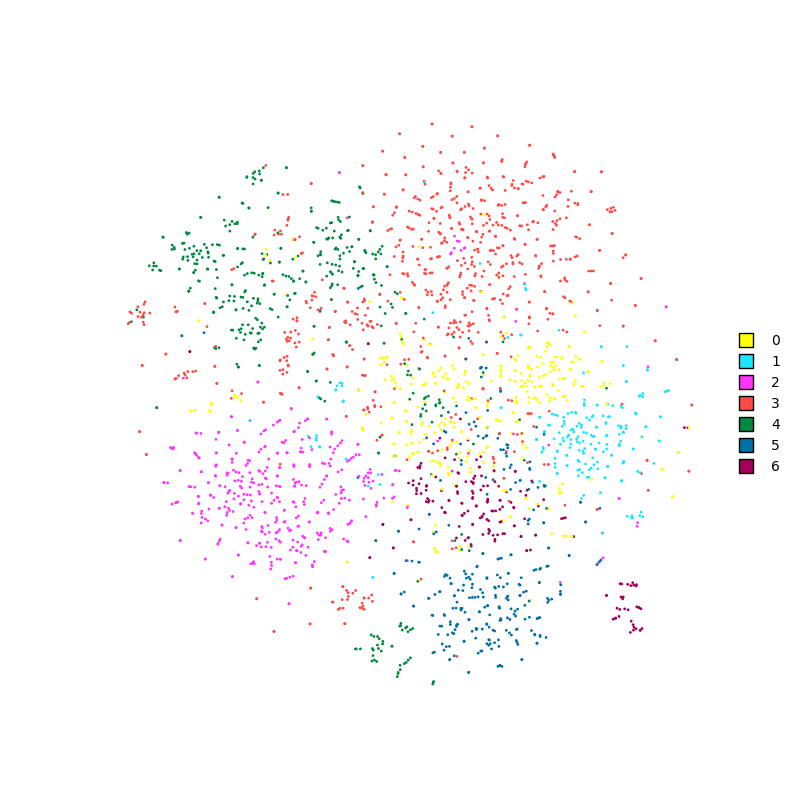}
}
\hspace{-5mm}
\subfigure [\texttt{Cora}-DGNN]{
\includegraphics[width=4.2cm]{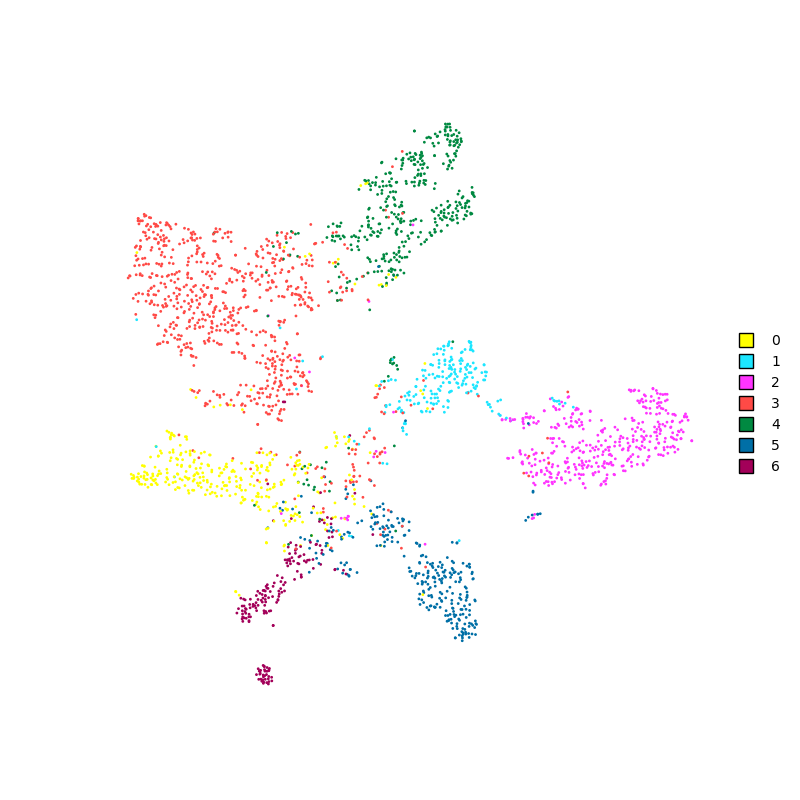}
}
\hspace{-5mm}
\subfigure [\texttt{Citeseer}-RAW]{
\includegraphics[width=4.2cm]{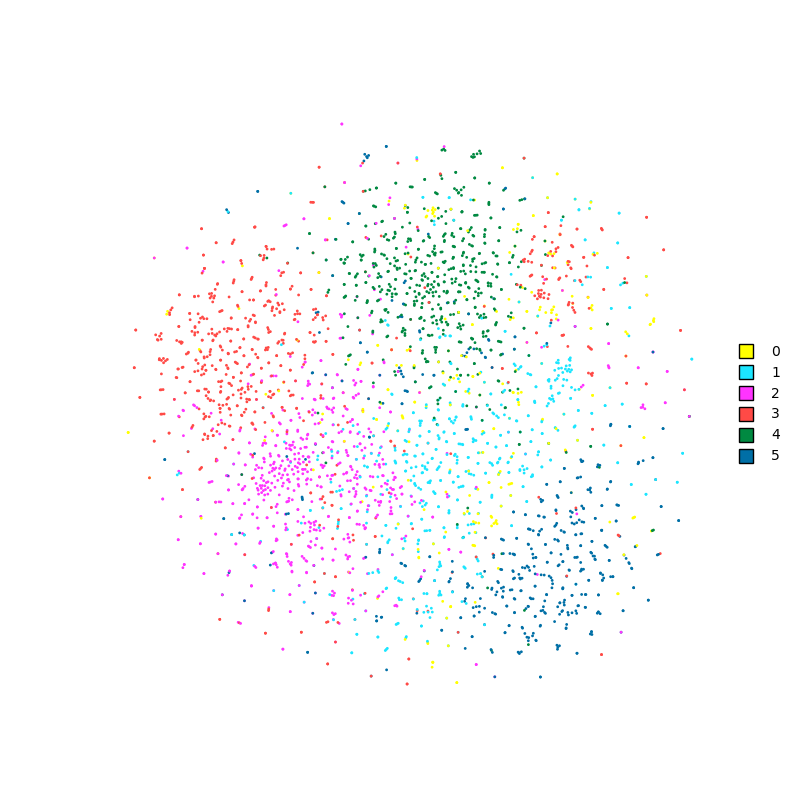}
}
\hspace{-5mm}
\subfigure [\texttt{Citeseer}-DGNN]{
\includegraphics[width=4.2cm]{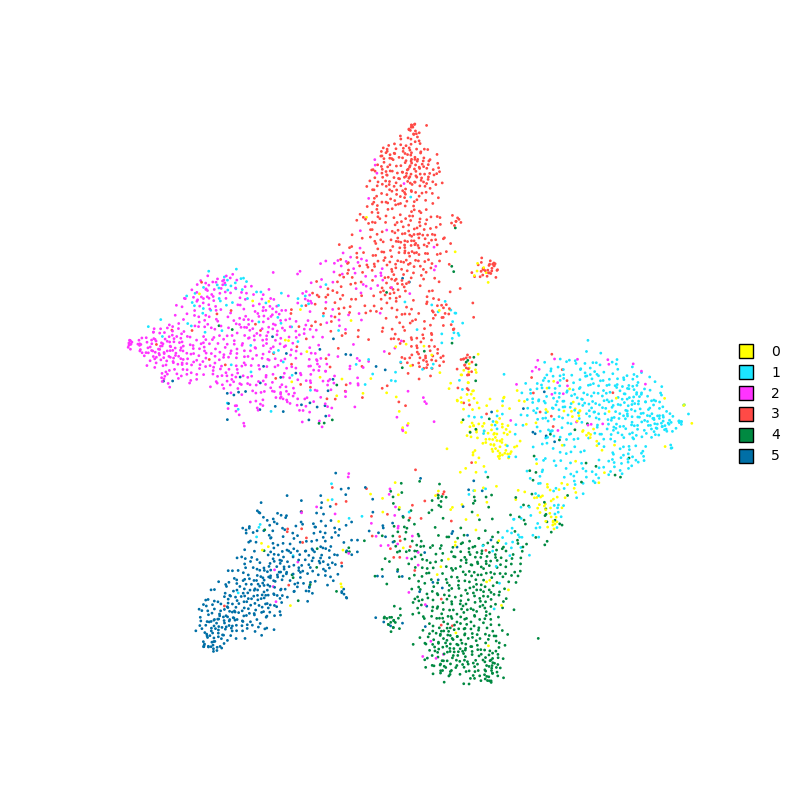}
}

\subfigure [\texttt{Chameleon}-RAW]{
\includegraphics[width=4.2cm]{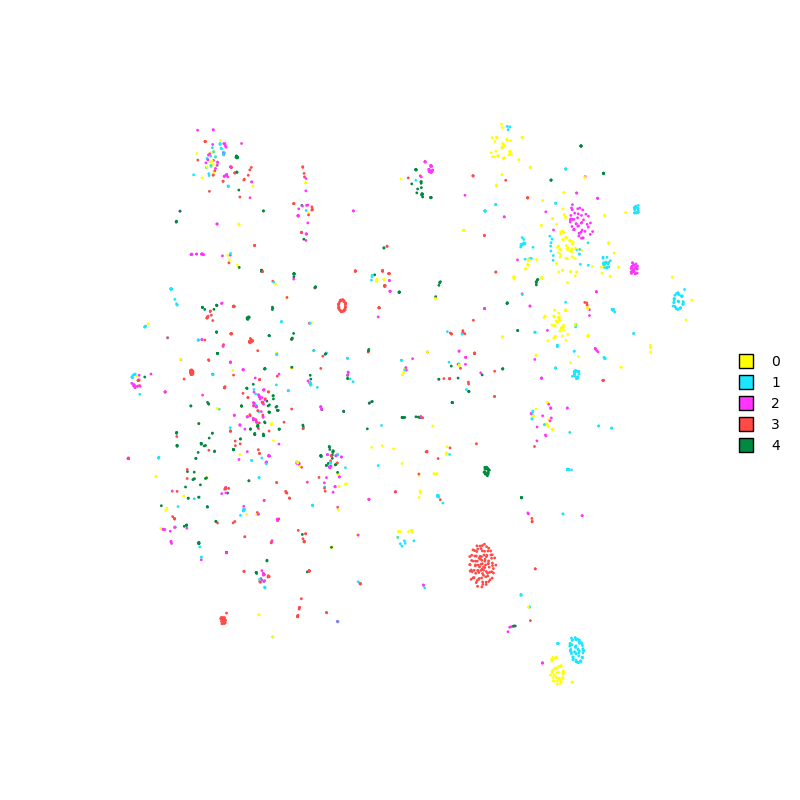}
}
\hspace{-5mm}
\subfigure [\texttt{Chameleon}-DGNN]{
\includegraphics[width=4.2cm]{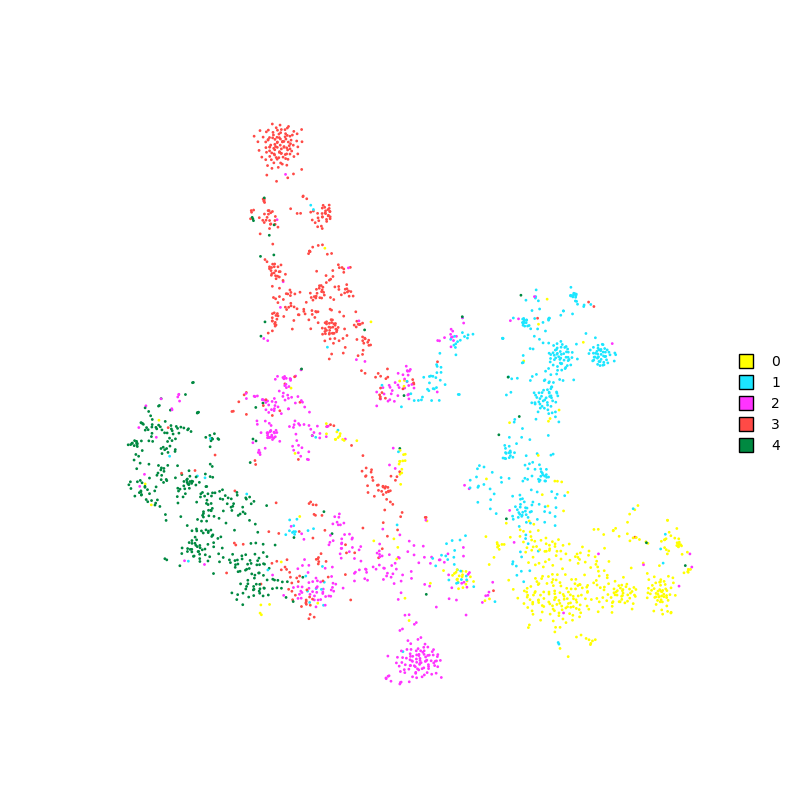}
}
\hspace{-5mm}
\subfigure [\texttt{Squirrel}-RAW]{
\includegraphics[width=4.2cm]{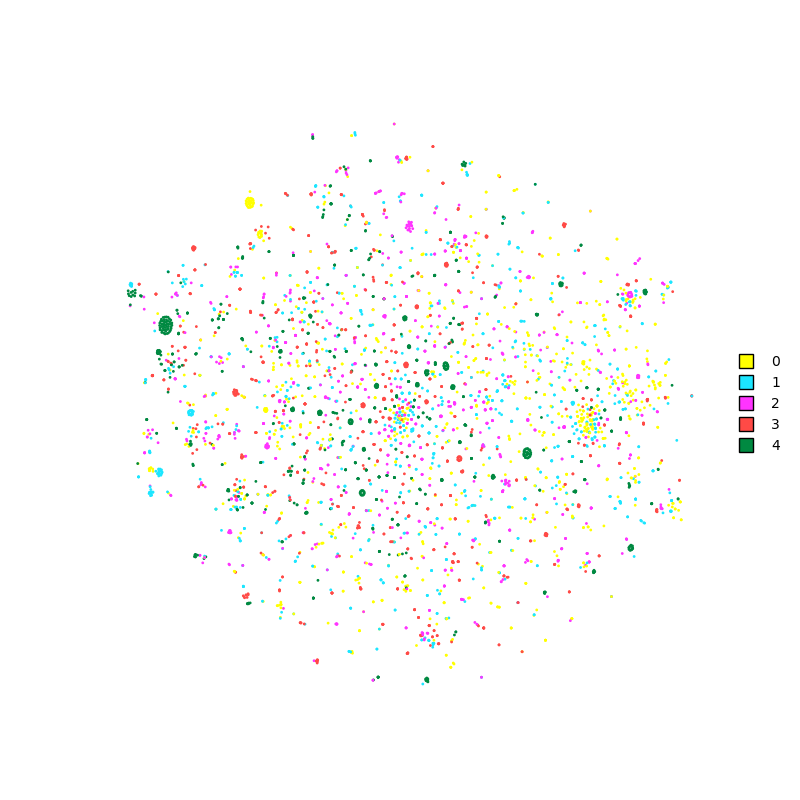}
}
\hspace{-5mm}
\subfigure [\texttt{Squirrel}-DGNN]{
\includegraphics[width=4.2cm]{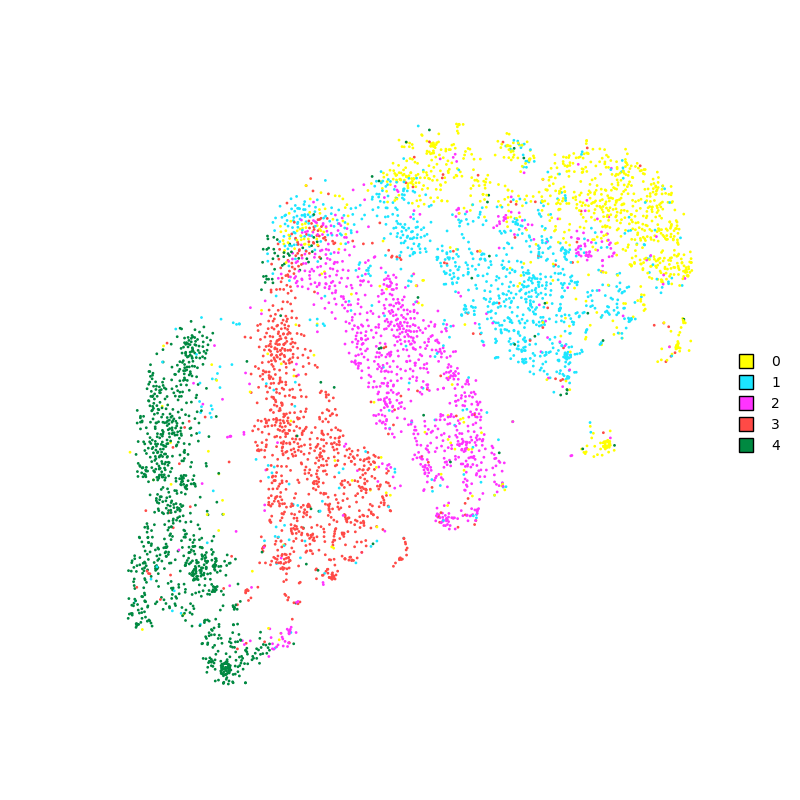}
}

\subfigure [\texttt{Computers}-RAW]{
\includegraphics[width=4.2cm]{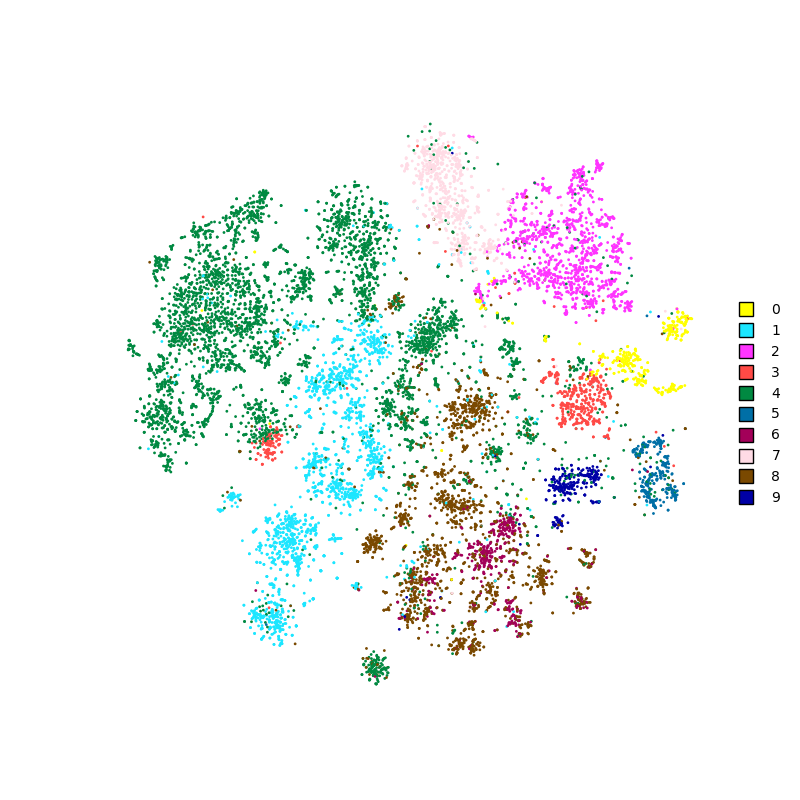}
}
\hspace{-5mm}
\subfigure [\texttt{Computers}-DGNN]{
\includegraphics[width=4.2cm]{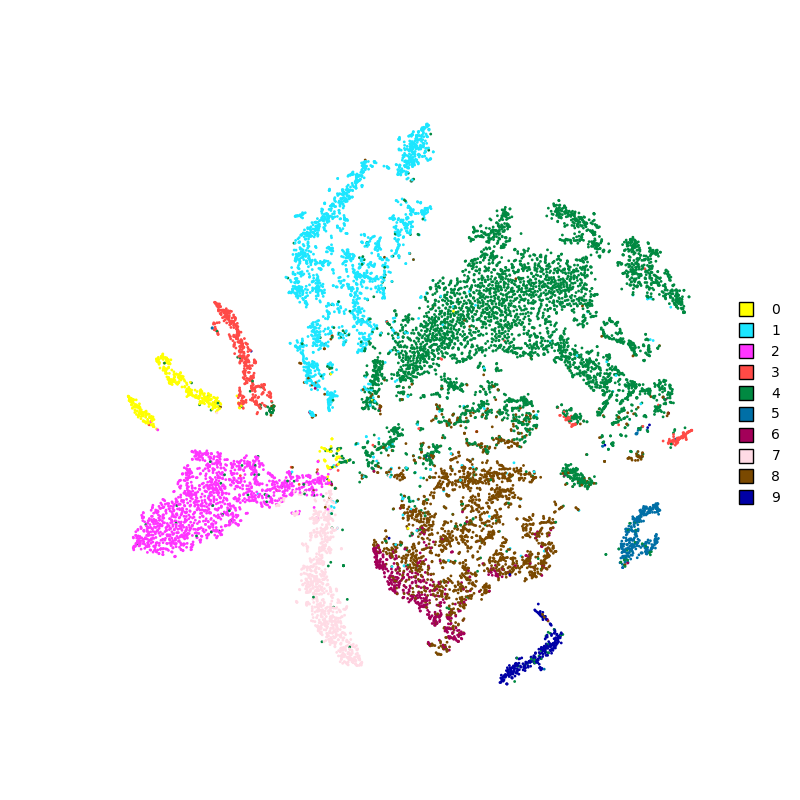}
}
\hspace{-5mm}
\subfigure [\texttt{Photo}-RAW]{
\includegraphics[width=4.2cm]{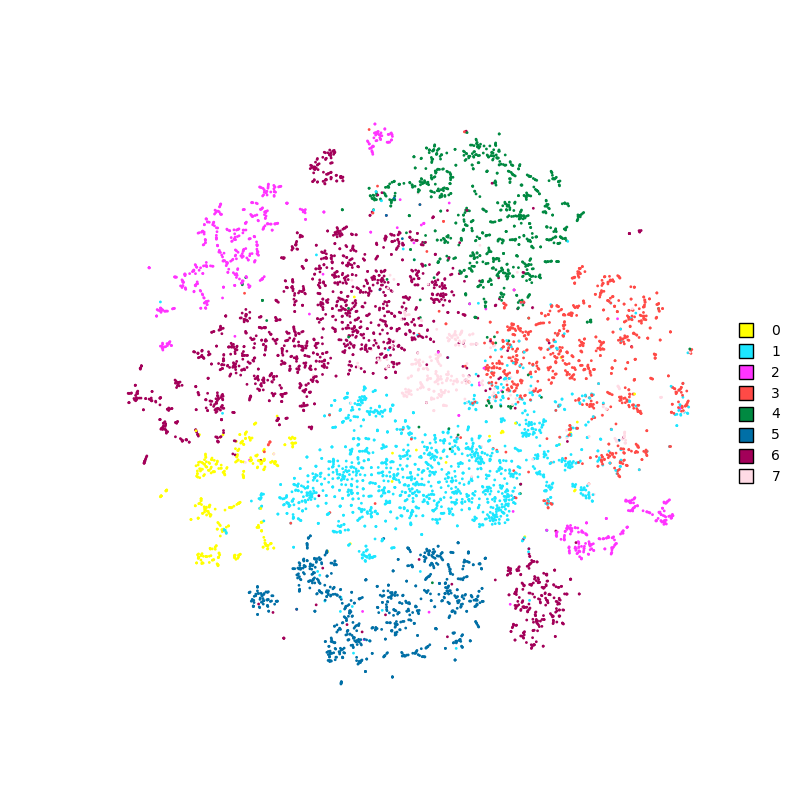}
}
\hspace{-5mm}
\subfigure [\texttt{Photo}-DGNN]{
\includegraphics[width=4.2cm]{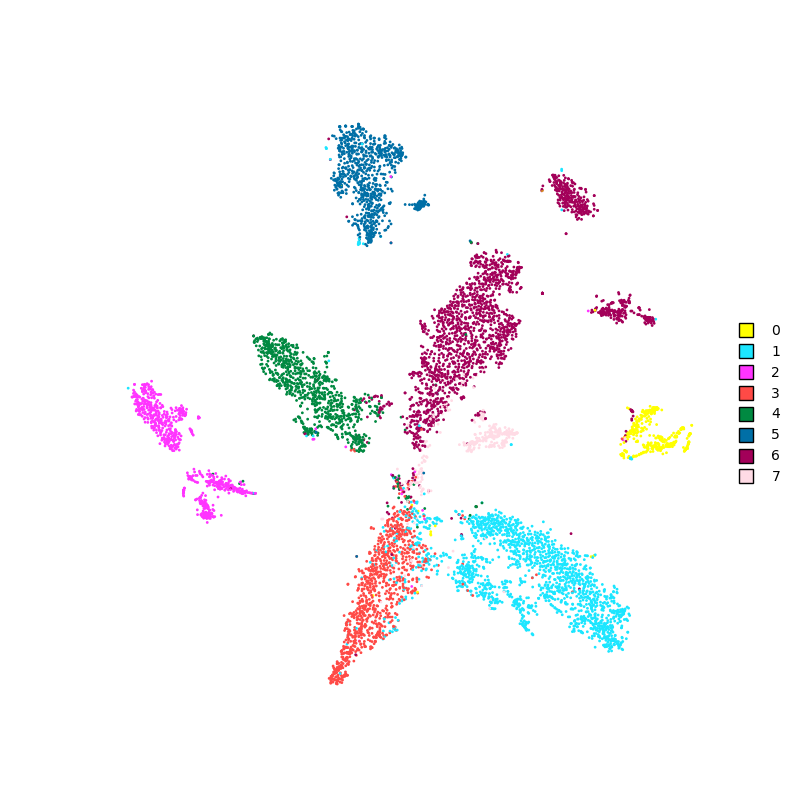}}
\caption{The t-SNE visualization for original attribute feature and node embedding representation obtained
by DGNN in a 2-D space on the all datasets.}\label{visualization}
\end{figure*}

To intuitively show the advantage of DGNN in discriminative embedding representation learning, we exhibit the original attribute feature and embedding representation learned by our DGNN  using t-SNE visualization technology. The visualization results are reported in Fig.~\ref{visualization}, where different colors are utilized to mark nodes from various classes. The distribution and shape reflect the discriminability of embedding representation learned by  DGNN. As illustrated in Fig.~\ref{visualization}, for original attribute feature, the clusters of different classes intersect and their shapes are irregular.
In contrast, the clusters of embedding representation learned by DGNN are almost disjoint, and the distribution of nodes from same class is compact. This demonstrates that DGNN is capable of improving the discriminative quality of embedding representation.

\begin{figure}
\centering
%\hspace{-10mm}
\subfigure [\texttt{Cora}]{
\includegraphics[width=4.35cm]{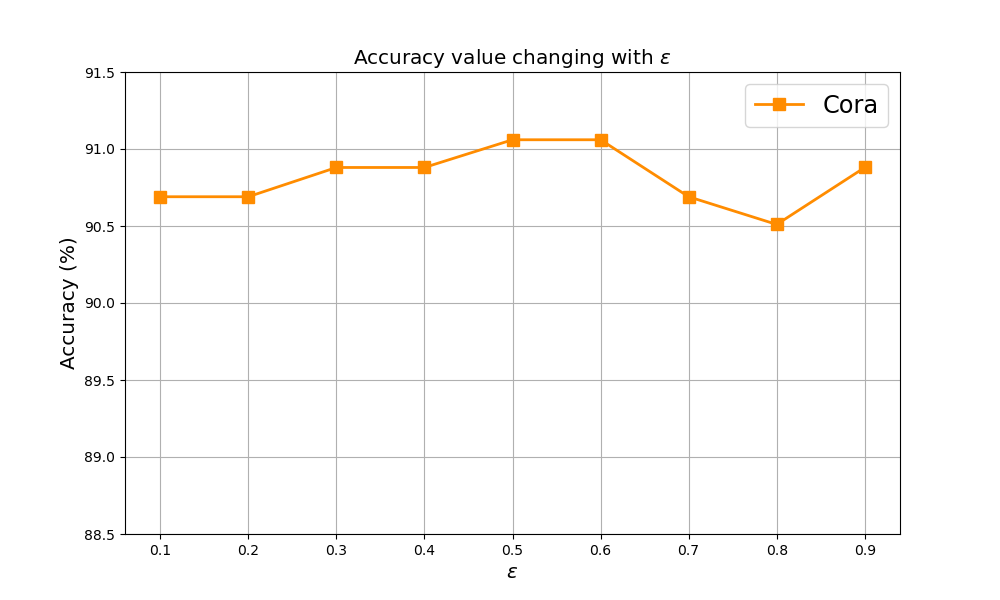}
}
\hspace{-6mm}
\subfigure [\texttt{Citeseer}]{
\includegraphics[width=4.35cm]{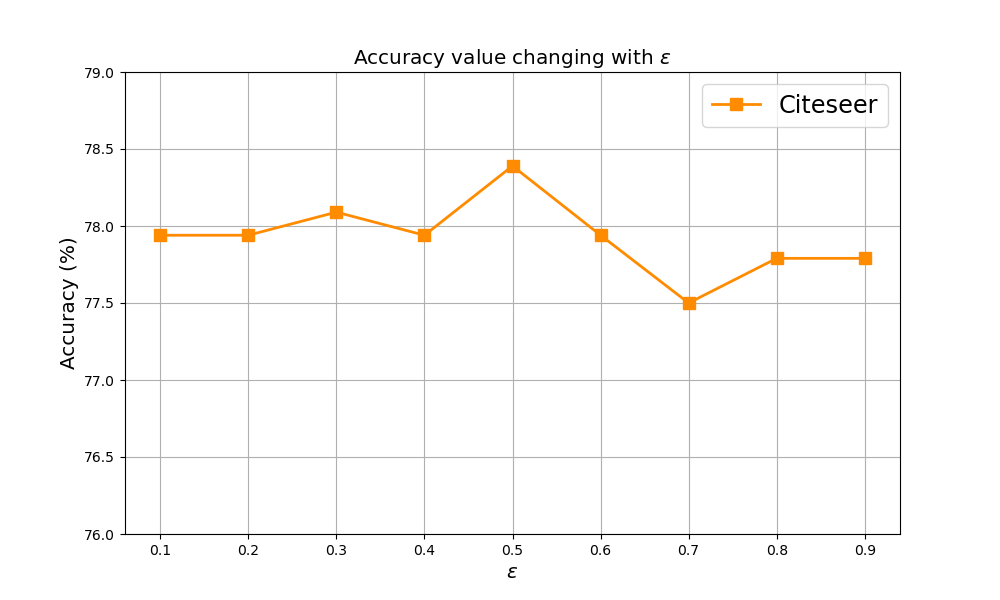}
}
\hspace{-12mm}
\subfigure [\texttt{Chameleon}]{
\includegraphics[width=4.35cm]{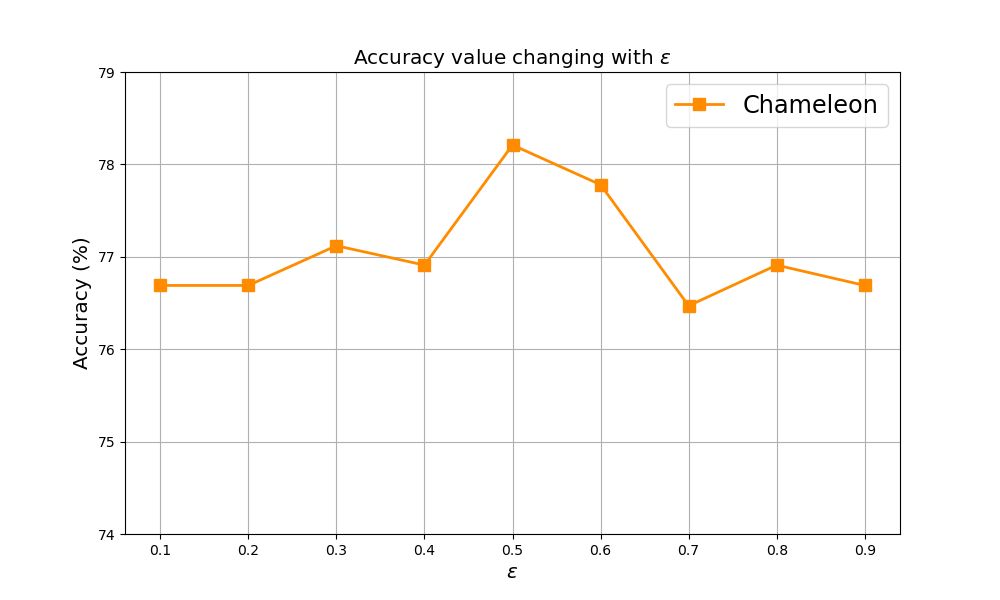}
}
\hspace{-6mm}
\subfigure [\texttt{Squirrel}]{
\includegraphics[width=4.35cm]{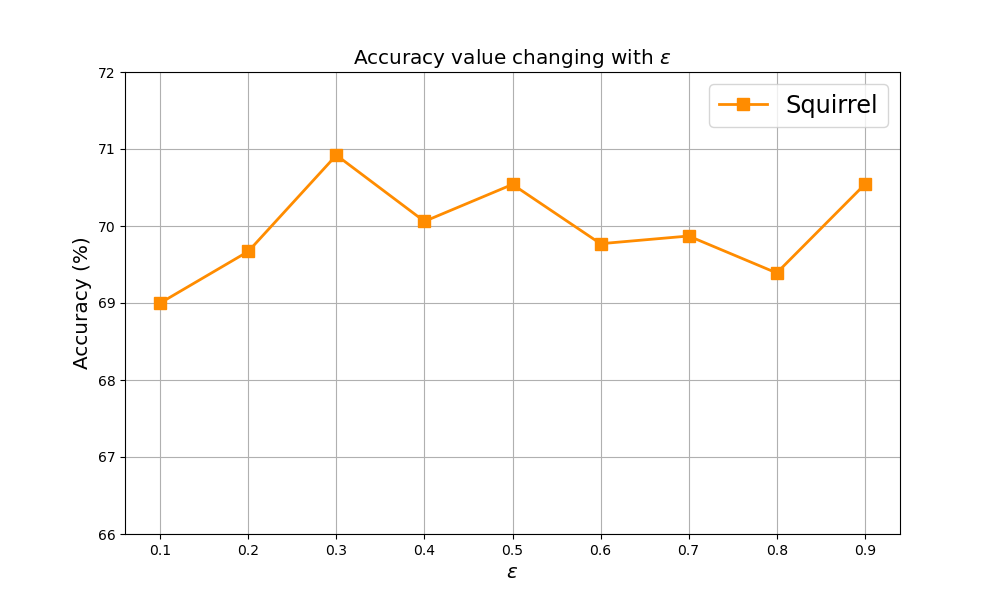}
}
\caption{The classification accuracy of DGNN with different $\varepsilon$ on the \texttt{Cora}, \texttt{Citeseer}, \texttt{Chameleon} and \texttt{Squirrel} datasets.}\label{epsilon_para}
\end{figure}

\begin{figure*}
\centering
%\hspace{-10mm}
\subfigure [\texttt{Cora}: $\beta=0.001$]{
\includegraphics[width=5cm]{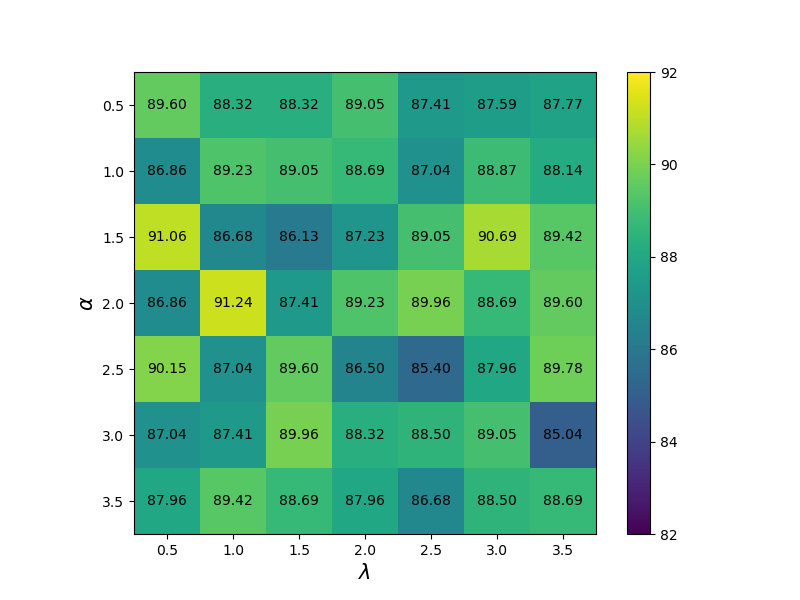}
}
\hspace{-12mm}
\subfigure [\texttt{Cora}: $\beta=0.005$]{
\includegraphics[width=5cm]{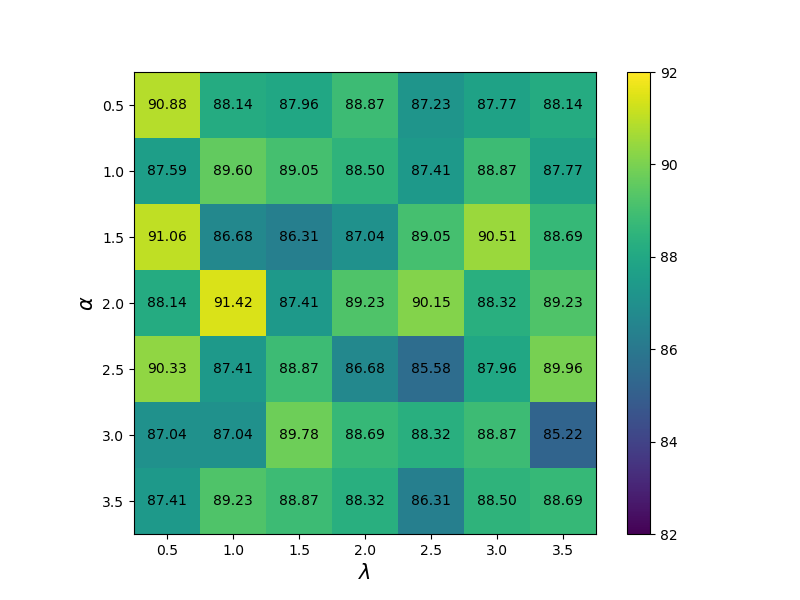}
}
\hspace{-12mm}
\subfigure [\texttt{Cora}: $\beta=0.01$]{
\includegraphics[width=5cm]{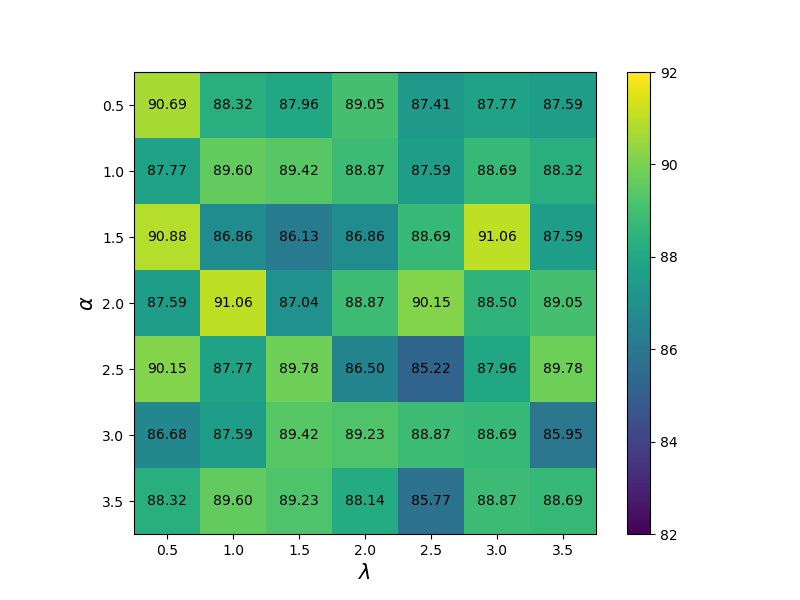}
}
\hspace{-12mm}
\subfigure [\texttt{Cora}: $\beta=0.02$]{
\includegraphics[width=5cm]{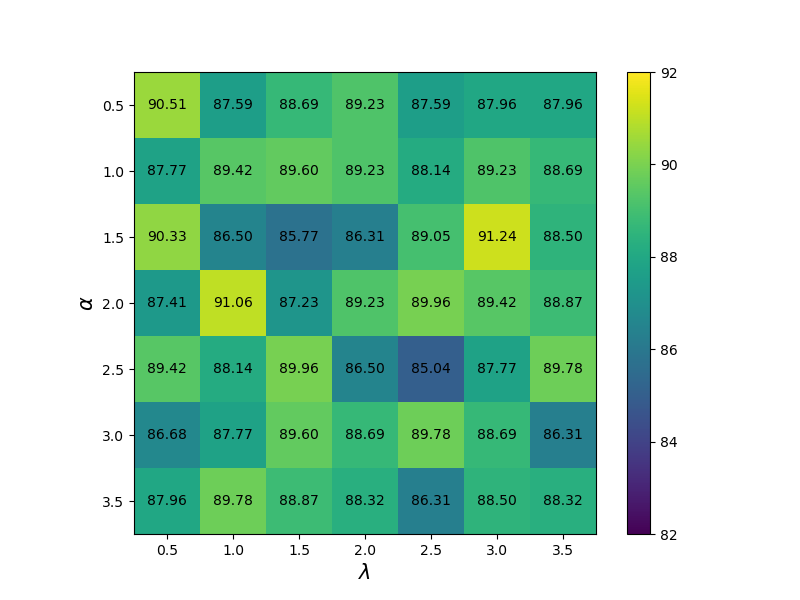}
}
\subfigure [\texttt{Squirrel}: $\beta=0.001$]{
\includegraphics[width=5cm]{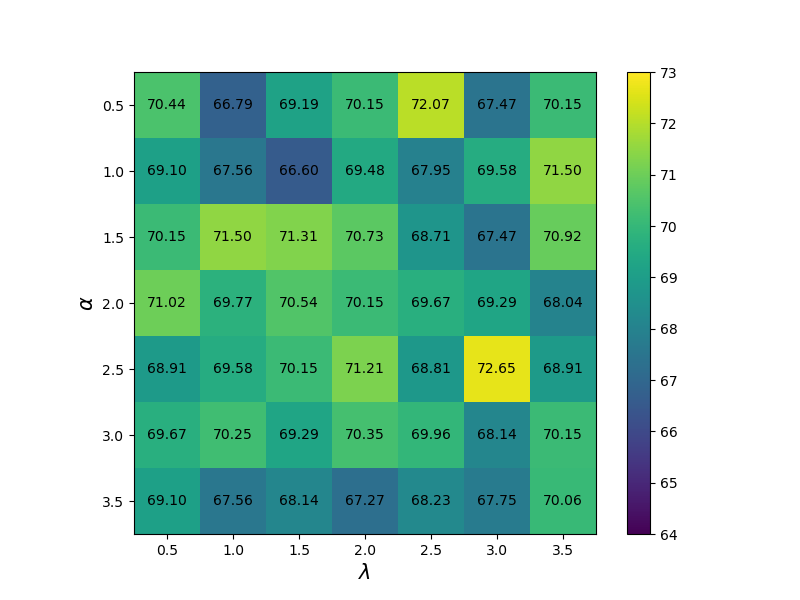}
}
\hspace{-12mm}
\subfigure [\texttt{Squirrel}: $\beta=0.005$]{
\includegraphics[width=5cm]{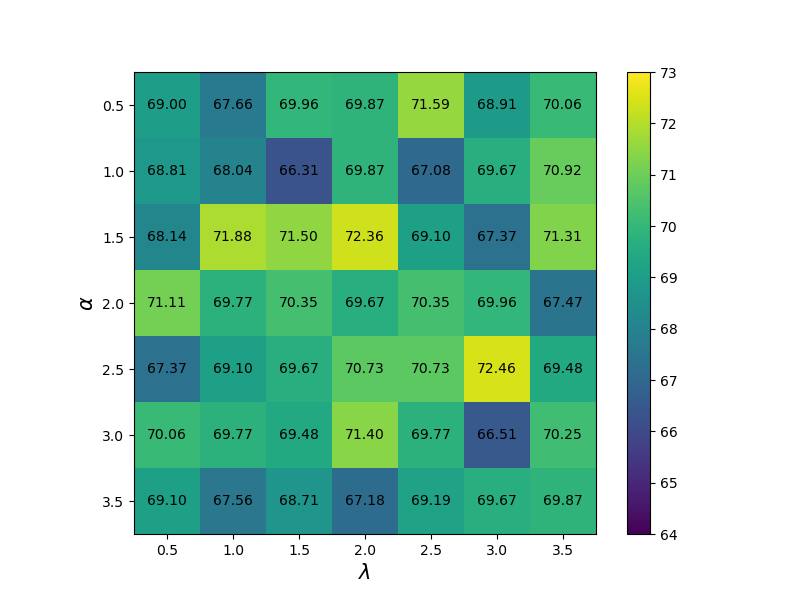}
}
\hspace{-12mm}
\subfigure [\texttt{Squirrel}: $\beta=0.01$]{
\includegraphics[width=5cm]{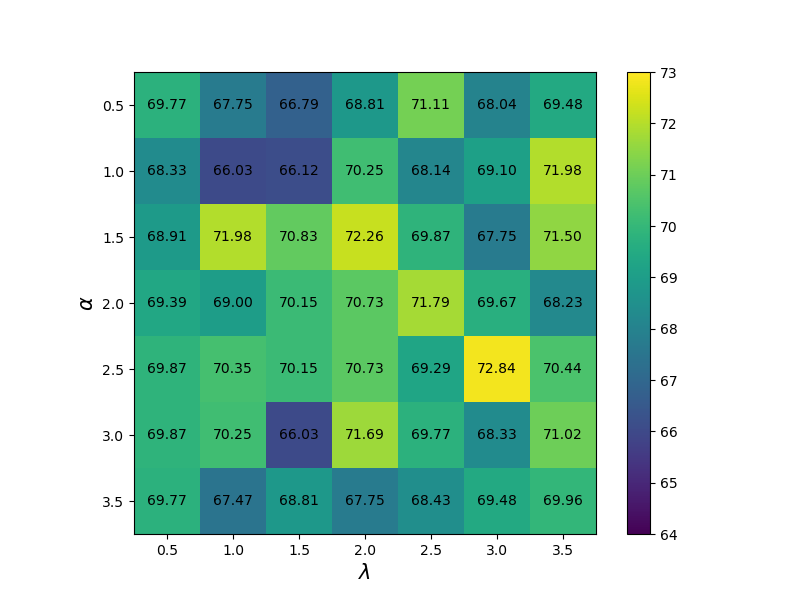}
}
\hspace{-12mm}
\subfigure [\texttt{Squirrel}: $\beta=0.02$]{
\includegraphics[width=5cm]{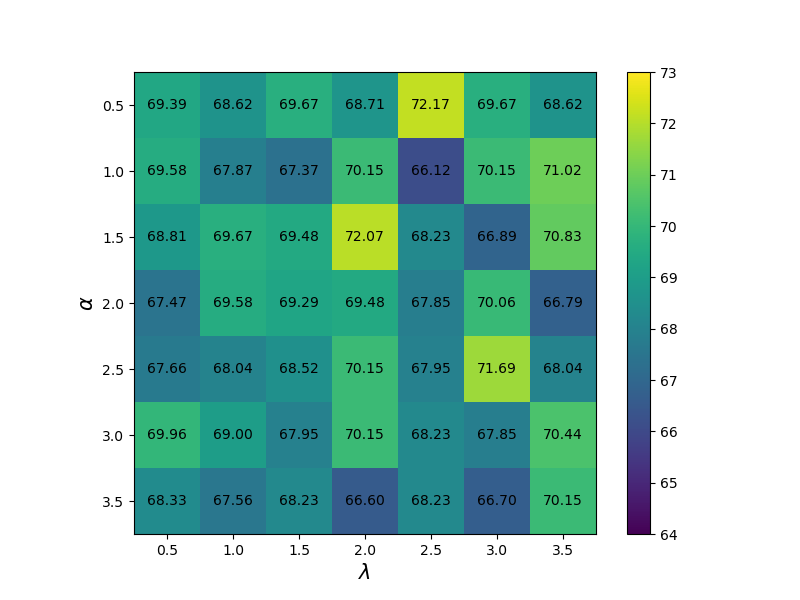}
}
\caption{The parameter sensitivity analysis of DGNN with respect to $\lambda$, $\alpha$, and $\beta$ on the \texttt{Cora} and \texttt{Squirrel} datasets.}\label{parameter analysis}
\end{figure*}

\subsection{Analysis of Balance Coefficient \texorpdfstring{$\varepsilon$}.}

In the previous experimental setting, we empirically set the balance coefficient $\varepsilon$ in structural consistency term to 0.5. Here, we adjust the coefficient $\varepsilon$ from $\{0.1, 0.2, \cdots, 0.9\}$ and examine its impact on classification accuracy in the Fig.~\ref{epsilon_para}. The classification accuracy with a coefficient of $\varepsilon=0.5$ is satisfactory and almost optimal,  confirming that the embedding representations of both topological graph and attribute graph are equally important for exploring structural consistency. And, the performance improvement of DGNN also depends on the mutual  enhancement of the three representations through the exploration of structural correlation.

\subsection{Analysis of Parameters \texorpdfstring{$\lambda$}., \texorpdfstring{$\alpha$}., and \texorpdfstring{$\beta$}.}\label{parameter_analysis}

Our DGNN has three hyper-parameters $\lambda$, $\alpha$, and $\beta$, , which regulate different terms in the objective function of our DGNN. The parameters $\lambda$ and $\alpha$ balance the contributions of the original topological graph the attribute graph, and are searched from the range $\{0.5, 1.0, 1.5, 2.0, 2.5, 3.0, 3.5\}$. The search set of $\beta$ is $\{0.001, 0.005, 0.01, 0.02\}$. The influences of these parameters on node classification performance for \texttt{Cora} and \texttt{Squirrel} datasets are shown in Fig.~\ref{parameter analysis}. Specifically, for the sake of visualization, we show the impact of $\lambda$ and $\alpha$ on classification accuracy when $\beta$ is fixed and selected from its corresponding range. The results indicate that the performance of DGNN can be significantly improved when appropriate parameters are utilized. DGNN exhibits relative robustness to the value of $\beta$, underscoring the effectiveness of the designed structural consistency strategy for exploring the consensus semantic correlation among various representations. Different from $\beta$, the performance of DGNN fluctuates slightly when the parameters $\lambda$ and $\alpha$ change. This behavior is attributed to the distinct roles that the original topological graph and attribute-based semantic graph play for various datasets. Although, both are essential for learning complementary and comprehensive embedding representations.

\begin{figure*}
\centering
%\hspace{-10mm}
\subfigure [\texttt{Citeseer}]{
\includegraphics[width=8cm]{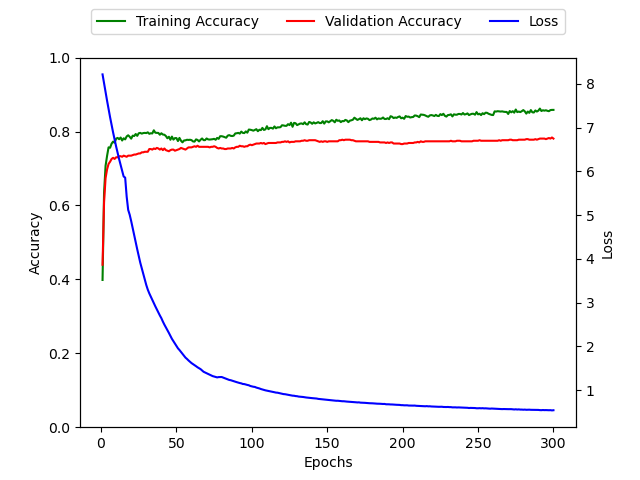}
}
\hspace{-1mm}
\subfigure [\texttt{Chameleon}]{
\includegraphics[width=8cm]{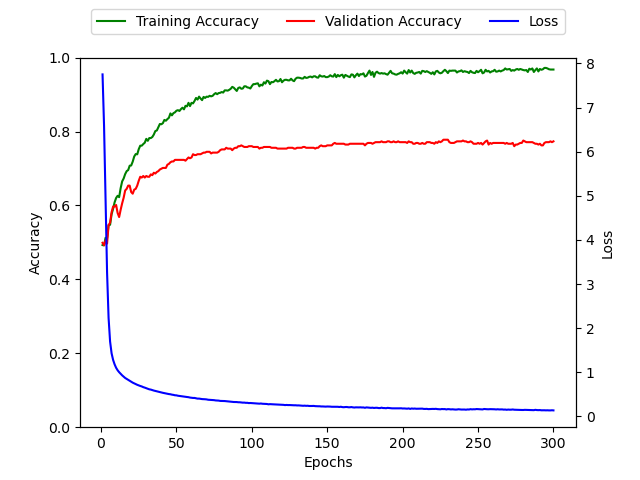}
}
\caption{The loss value and classification accuracy versus the training epochs of DGNN on the \texttt{Citeseer} and \texttt{Chameleon} datasets.}\label{convergence}
\end{figure*}

\subsection{Convergence Analysis}

We experimentally show the convergence of the DGNN method. Specifically, we draw the value of objective function along with classification accuracy (including training accuracy and validation accuracy) versus the training epochs on the \texttt{Citeseer} and \texttt{Chameleon} datasets. As shown in Fig.~\ref{convergence}, the value of objective function undergoes a sharp decrease and then stabilizes. In addition, the classification accuracy of training set and validation set is increasing overall and achieves a relatively stable value. These experimental results obviously verify the strong stability and convergence of our DGNN method.

\section{Conclusion}\label{Sec:6}
In this paper, we propose a novel optimization framework induced graph convolutional network
dubbed DGNN, which learns distinctive embedding representations for attribute feature and graph structures in a decoupled manner. The DGNN adeptly mitigates interference at feature space between node attributes and graph, resulting in comprehensive node embedding representations  that contain complementary information. Further, the structural consistency constraint, incorporating a shared reconstruction factor, is deigned to reduce the redundancy and establish consensus semantic correlation  across various representations by promoting consistent reconstruction adjacency relationships. The abundant experimental results conducted on node classification task verify the effectiveness and superiority of proposed DGNN method. Although, similar to general GNNs methods, the computational complexity of DGNN scales quadratically with respect to the number of nodes, which limits its scalability to large-scale graph data. Exploring the complexity-economic version for DGNN is an interesting topic of future work.

\section*{Acknowledgements}
This research was supported by the
National Natural Science Foundation of China under Grant 62172023; in part by the Beijing Natural Science Foundation under Grant 4222021 and 4244085; in part by the Postdoctoral Fellowship Program of CPSF under Grant GZC20230203; in part by
the China Postdoctoral Science Foundation under Grant 2023M740201.

\bibliographystyle{IEEEtran}
\bibliography{DGNN}

\begin{IEEEbiography}[{\includegraphics[width=1in,height=1.25in,clip,keepaspectratio]{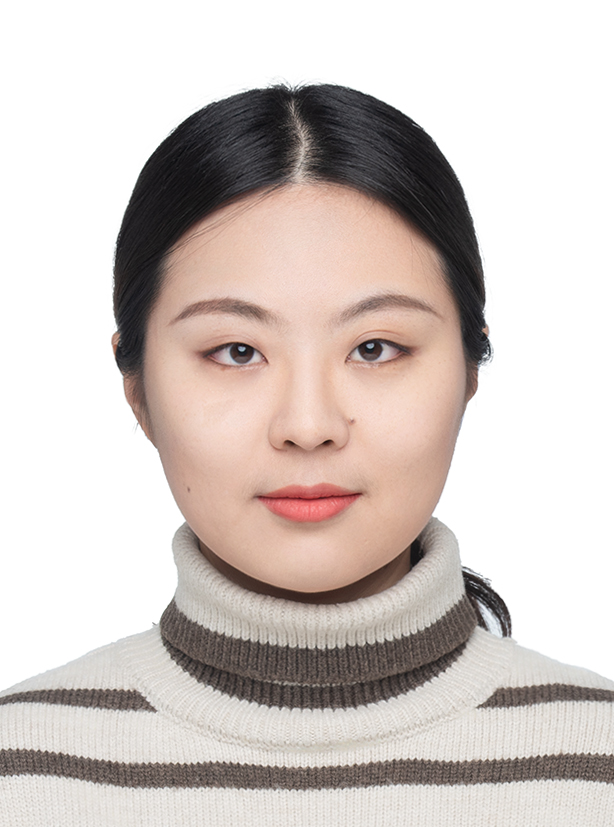}}]
{Jinlu Wang} received the B.S. degree from the
YanShan University, Hebei,
China, in 2017. She is currently pursuing the Ph.D.
degree with the Faculty of Information Technology, Beijing University of Technology, Beijing. Her
current research interests include machine learning, graph representation learning and pattern recognition.
\end{IEEEbiography}
\vspace{-0.8cm}

\begin{IEEEbiography}[{\includegraphics[width=1in,height=1.25in,clip,keepaspectratio]{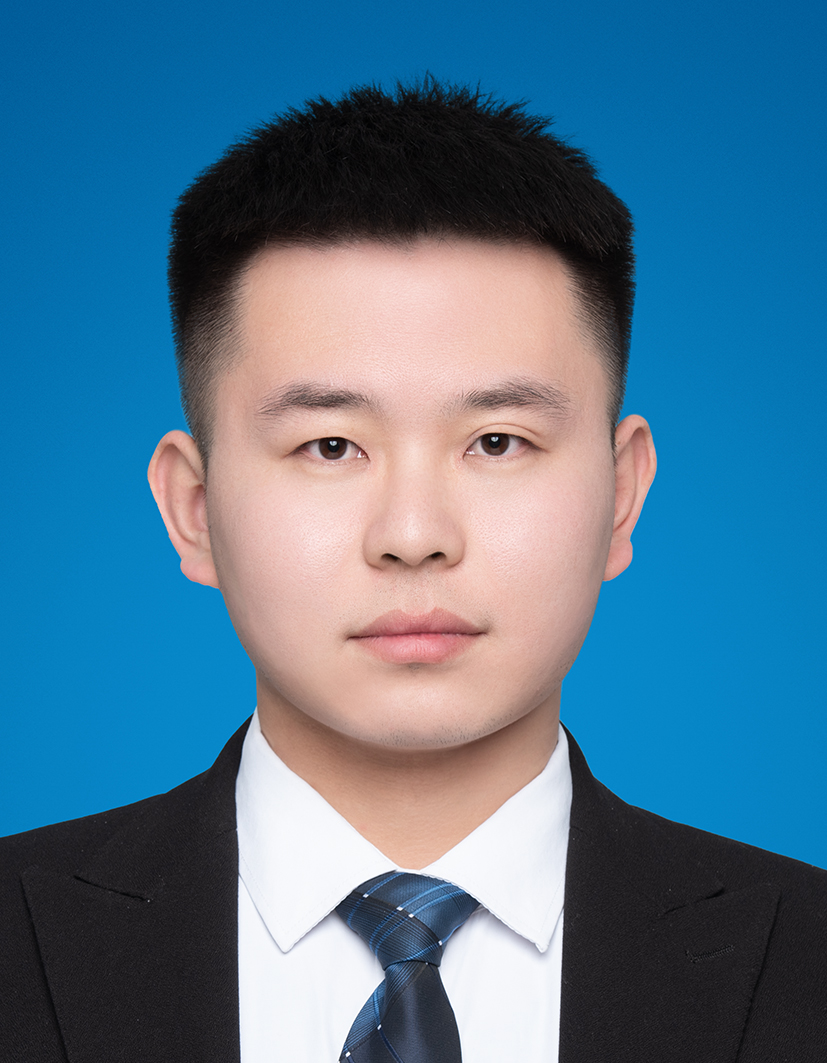}}]
{Jipeng Guo} received the Ph.D. degree in control science and engineering from
the Beijing University of Technology, Beijing,
China, in 2023. 

He is currently a Lecturer with the College of Information Science and Technology, Beijing University of Chemical Technology, Beijing 100029, China. He has published several papers in top 
journals and conferences, such as IEEE TPAMI, IEEE TIP, IEEE TNNLS, ACM TKDD, Neural Networks, AAAI, and ICASSP. His current research interests include pattern recognition, machine learning, graph representation learning and
clustering method.
\end{IEEEbiography}
\vspace{-0.8cm}

\begin{IEEEbiography}[{\includegraphics[width=1in,height=1.25in,clip,keepaspectratio]{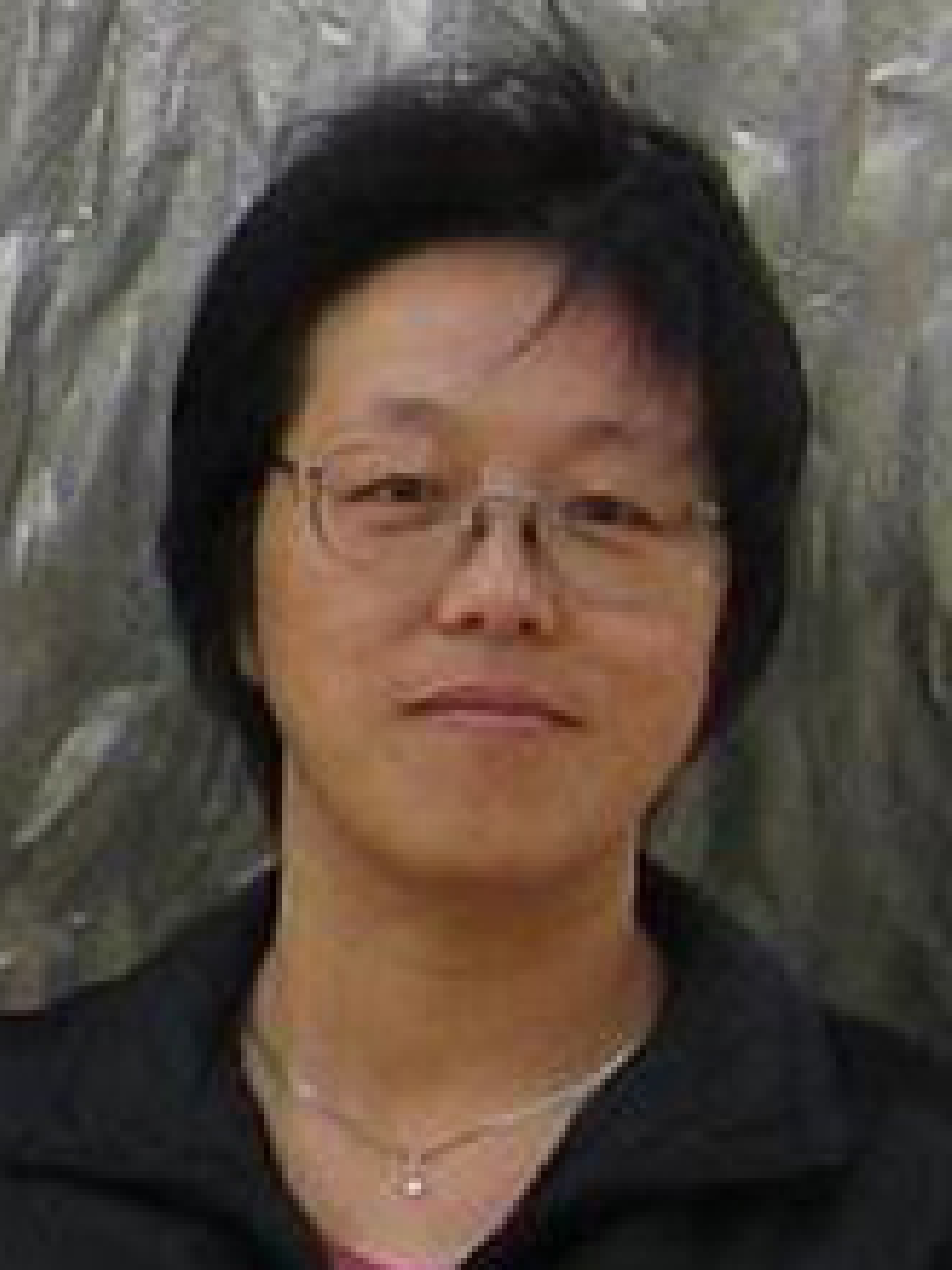}}]
{Yanfeng Sun}received her Ph.D. degree from Dalian University of Technology in 1993. 

She is currently a Professor with the Faculty of Information Technology, Beijing university of Technology, Beijing, China. She is also a researcher with the Beijing Key Laboratory of Multimedia and Intelligent Software Technology, and with the Beijing Advanced Innovation Center for Future Internet Technology. She is the membership of China Computer Federation. She has published over 150 conferences and journals include IEEE TPAMI/TIP/TNNLS/TCYB/TCSVT, CVPR, IJCAI, AAAI. 
Her research interests are machine learning and image processing.
\end{IEEEbiography}
\vspace{-0.4cm}

\begin{IEEEbiography}[{\includegraphics[width=1in,height=1.25in,clip,keepaspectratio]{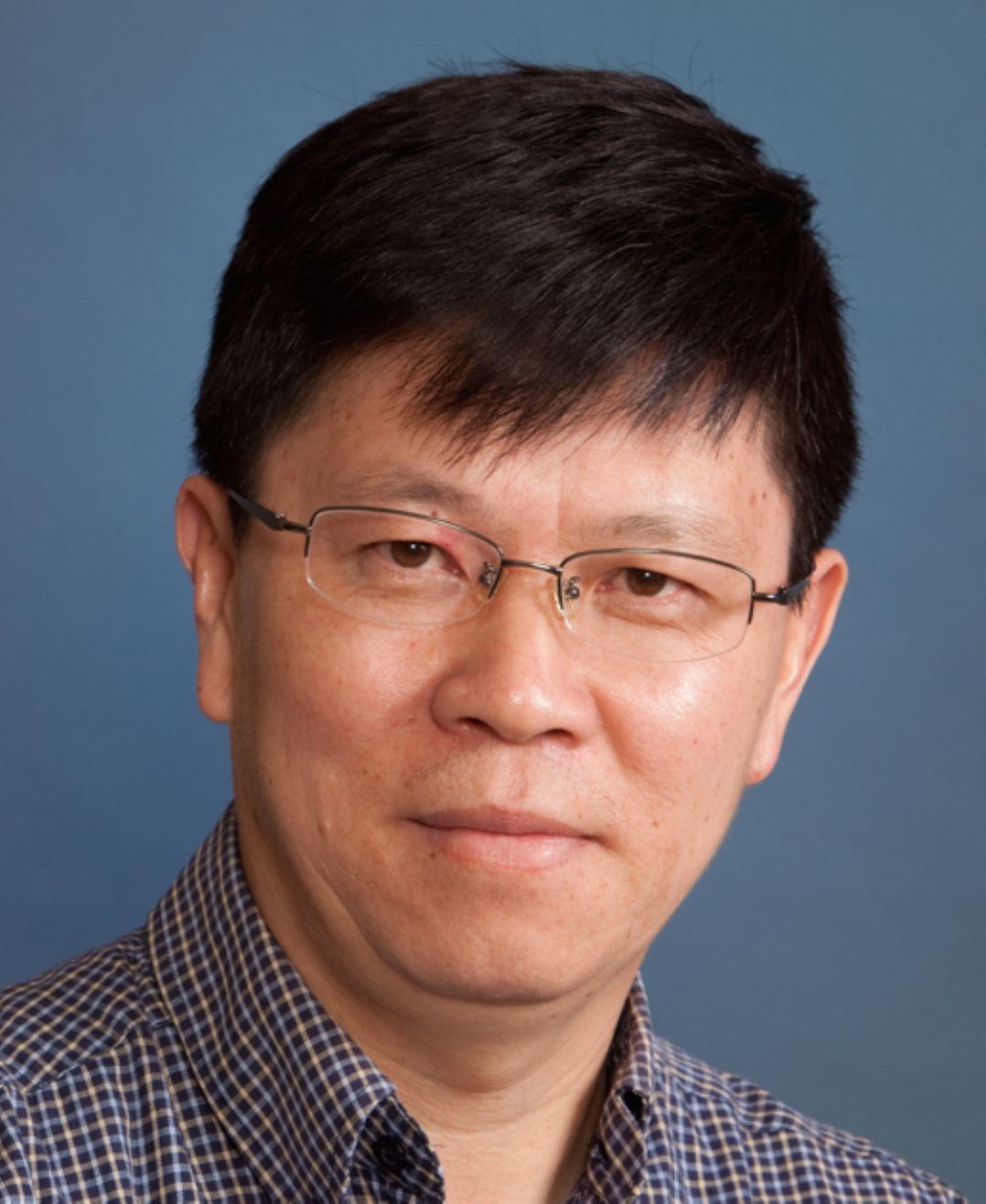}}]
{Junbin Gao} graduated from Huazhong University of Science and Technology (HUST),
China in 1982 with a BSc. in Computational Mathematics and
obtained his PhD. from Dalian University of Technology, China in 1991.

He is a Professor of Big Data Analytics in the University of Sydney Business School at the University of Sydney and was a Professor in Computer Science
in the School of Computing and Mathematics at Charles Sturt
University, Australia. He was a senior lecturer, a lecturer in Computer Science from 2001 to 2005 at the
University of New England, Australia. From 1982 to 2001 he was an
associate lecturer, lecturer, associate professor, and professor in
Department of Mathematics at HUST. His main research interests
include machine learning, data analytics, Bayesian learning and
inference, and image analysis.
\end{IEEEbiography}
\vspace{-0.4cm}

\begin{IEEEbiography}[{\includegraphics[width=1in,height=1.25in,clip,keepaspectratio]{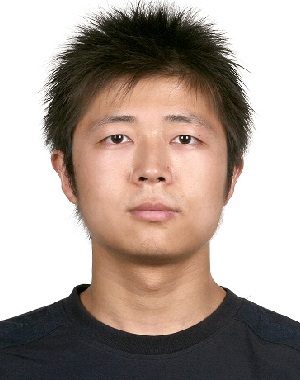}}]
{Shaofan Wang}(Member, IEEE) received the B.S. and Ph.D. degrees in computational mathematics from the Dalian University of Technology, Dalian,
China, in 2003 and 2010, respectively. He is an Associate Professor with the Beijing Key Laboratory of Multimedia and Intelligent Software Technology, Faculty of Information Technology, Beijing University of Technology, Beijing, China.
His research interests include pattern recognition and machine learning.
\end{IEEEbiography}
\vspace{-0.4cm}

\begin{IEEEbiography}[{\includegraphics[width=1in,height=1.25in,clip,keepaspectratio]{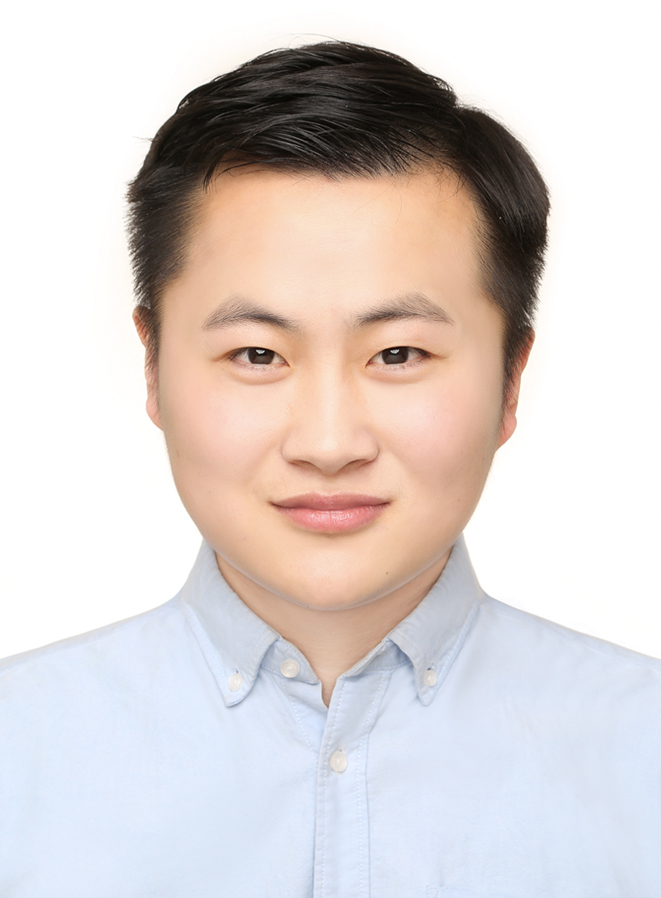}}]
{Yachao Yang} received the B.S. degree from
the Shandong University of Technology, Shandong,
China, in 2019. He is currently pursuing the Ph.D.
degree with the Faculty of Information Technology, Beijing University of Technology, Beijing. He has published several papers in top journals and conferences, such as AAAI, TNNLS, Neural Networks and Information Sciences.
His current research interests include computer
vision, pattern recognition, machine learning, and
clustering method.
\end{IEEEbiography}
\vspace{-0.4cm}

\begin{IEEEbiography}[{\includegraphics[width=1in,height=1.25in,clip,keepaspectratio]{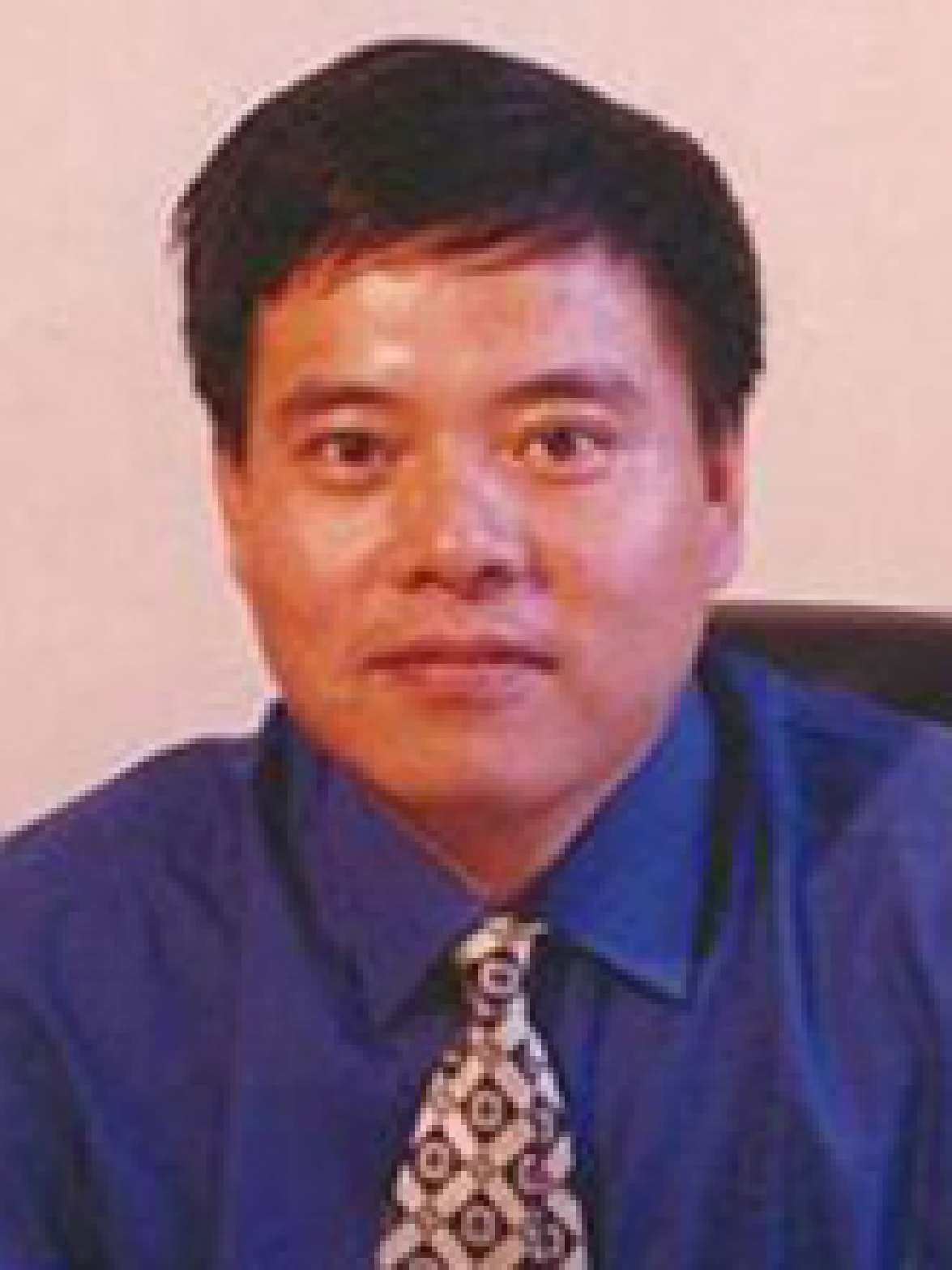}}]
{Baocai Yin} received his Ph.D. from Dalian University of Technology in 1993.

He is currently a Professor with the Faculty of Information Technology, Beijing university of Technology. He is also a researcher with the Beijing Key Laboratory of Multimedia and Intelligent Software Technology.
He is a member of the China Computer Federation. He has authored or coauthored
more than 200 academic articles in prestigious international journals and conferences, including the IEEE TPAMI/TIP/TMM/TCYB/TNNLS/TCSVT, ICCV, CVPR, ECCV, AAAI, IJCAI, SIGGRAPH. His research interests include multimedia, image processing,
computer vision, and pattern recognition.
\end{IEEEbiography}
\vfill

\end{document}